%% file: acl_latex.tex
\pdfoutput=1
\documentclass[11pt]{article}

\usepackage[final]{acl}

\usepackage{times}
\usepackage{latexsym}

\usepackage[T1]{fontenc}
\usepackage[utf8]{inputenc}

\usepackage{microtype}

\usepackage{inconsolata}

\usepackage{graphicx}

\usepackage{hyperref}
\usepackage{url}
\usepackage{tabu}
\usepackage{booktabs}
\usepackage{multirow} 
\usepackage{xcolor}
\usepackage{wrapfig} 
\usepackage{colortbl}
\usepackage[most]{tcolorbox}
\usepackage{subcaption}
\usepackage{titletoc}
\usepackage{amssymb}
\usepackage{enumitem}
\usepackage[capitalize]{cleveref} 
\definecolor{wkyellow}{RGB}{255,241,177}
\definecolor{lightblue}{HTML}{CCE5FF}
\definecolor{lightpink}{HTML}{FFCCCC}

\definecolor{lightgray}{gray}{0.9}
\definecolor{goodblue}{HTML}{0071bc}
\definecolor{mybrown}{HTML}{804F46}
\definecolor{mygreen}{HTML}{50C92A}
\definecolor{myred}{HTML}{BD2A35}

\usepackage{CJKutf8}         % 支持中文、日文、韩文

\usepackage{pgfplots} % for colormap
\pgfplotsset{compat=1.18}

\tcbuselibrary{most,skins,theorems,breakable}
\tcbset{
  aibox/.style={
    width=\linewidth,
    top=7pt,
    bottom=2pt,
    colback=blue!6!white,
    colframe=black,
    colbacktitle=black,
    enhanced,
    center,
    breakable,
    attach boxed title to top left={yshift=-0.1in,xshift=0.15in},
    boxed title style={boxrule=0pt,colframe=white,},
  }
}
\newtcolorbox{AIbox}[2][]{aibox,title=#2,#1}

\title{How Well Does Mathematical Reasoning Transfer Across Languages in English-Centric LLMs?}

\author{Wen Yang\textsuperscript{1,2}, Junhong Wu\textsuperscript{1,2}, Chong Li\textsuperscript{1,2}, Chengqing Zong\textsuperscript{1,2}, Jiajun Zhang\textsuperscript{1,2,3} \thanks{\ \ Corresponding author} \\
        \\
        \textsuperscript{1}~School of Artificial Intelligence, University of Chinese Academy of Sciences\\
        \textsuperscript{2}~Institute of Automation, Chinese Academy of Sciences \\
        \textsuperscript{3}~Wuhan AI Research \\
        \texttt{\{yangwen2023, wujunhong2021, lichong2021\}@ia.ac.cn} \\
        \texttt{\{cqzong, jjzhang\}@nlpr.ia.ac.cn}
        }

\begin{document}
\maketitle
\begin{abstract}
% Recent advances in Reinforcement Post-Training (RPT) have substantially improved the mathematical reasoning capabilities of large language models (LLMs), but it remains unclear how well such gains transfer across languages. In this work, we study the cross-lingual transfer of mathematical reasoning in English-centric LLMs under controlled multilingual evaluation. We first evaluate a range of English-centric reasoning models on multilingual mathematical reasoning benchmarks, and the results show that cross-lingual transfer is highly variable across initial model choice, target language, and training paradigm. Through controlled comparative experiments, we further find that stronger English-centric initialization does not necessarily lead to stronger relative transfer efficiency, even when it yields better multilingual reasoning accuracy. Finally, we show that the largest improvement comes from moving beyond English-only supervision: adding a single parallel language yields a substantial gain, while further gains from additional languages are smaller but consistent. Overall, our results suggest that multilingual mathematical reasoning should be evaluated directly rather than inferred from English benchmarks alone.

Recent advances in Reinforcement Post-Training (RPT) have substantially improved the mathematical reasoning capabilities of large language models (LLMs), but it remains unclear how well such gains transfer across languages. In this work, we study cross-lingual mathematical reasoning transfer in English-centric LLMs under controlled multilingual evaluation. We first evaluate a range of English-centric reasoning models on multilingual mathematical reasoning benchmarks and find that transfer is highly variable across initial model choice, target language, and training paradigm. Through controlled comparative experiments, we further find that stronger English-centric initialization does not necessarily lead to stronger relative transfer efficiency, even when it yields higher multilingual reasoning accuracy. Finally, we show that the largest improvement comes from moving beyond English-only supervision: adding a single parallel language yields a substantial gain, while further gains from additional languages are smaller but consistent. Overall, our results highlight the need to evaluate multilingual mathematical reasoning directly, using both absolute performance and relative transfer-efficiency metrics.

\end{abstract}

\section{Introduction}

Recent advances in Reinforcement Post-Training (RPT)~\citep{jaech2024openai,team2025kimi,yang2025qwen3} have substantially improved the reasoning capabilities of large language models (LLMs), especially on mathematical benchmarks. In particular, Reinforcement Learning with Verifiable Rewards (RLVR)~\citep{lambert2024tulu, guo2025deepseek} has led to strong gains on datasets such as MATH~\citep{hendrycks2021measuring} and AIME. As these models continue to improve on English reasoning benchmarks, an important question is whether such gains transfer beyond English. Prior work has mainly studied transfer across tasks or modalities~\citep{chu2025sft, liu2025x, hu2026breaking, huan2025does, zhou2025does}, while cross-lingual transfer remains less understood.

This question matters for both multilingual use and model analysis. Strong English reasoning performance does not guarantee equally strong performance for multilingual users, especially when users require both correct answers and interpretable reasoning traces in their own language. More broadly, cross-lingual transfer offers a way to probe what English-centric post-training learns: transferable reasoning behavior, language-specific patterns, or some combination of both. This motivates the central question of this paper:
\begin{tcolorbox}[before skip=0.2cm, after skip=0.2cm, boxsep=0.0cm, middle=0.1cm, top=0.1cm, bottom=0.1cm]
\textit{\textbf{(Q)}} \textit{To what extent does mathematical reasoning acquired through English-centric post-training transfer across languages under controlled multilingual evaluation?}
\end{tcolorbox}

We answer this question in three stages. We first conduct an observational study over open-source English-centric reasoning models and find that cross-lingual transfer is highly variable across initial model choice, target language, and training paradigm. We then run controlled comparative studies on initialization type, model family, and model size. Across these settings, stronger English-centric initialization does not necessarily correspond to larger relative cross-lingual transfer gains. Finally, we study parallel multilingual supervision and find that it substantially improves transfer, with the largest gain appearing when moving from monolingual to bilingual training.

Our contributions are threefold. First, we provide an exploratory empirical study of cross-lingual mathematical reasoning transfer in English-centric LLMs, using both absolute performance and relative transfer-efficiency metrics. Second, under controlled GRPO-based post-training, we show that transfer varies across initialization type, model family, and model size. Third, we find that parallel multilingual supervision improves transfer in our setting, with the largest observed single-step gain from English-only to bilingual training and smaller but consistently positive gains from additional languages.

We organize the paper around the following research questions:
\begin{itemize}[left=0pt]
    \item \textit{\textbf{RQ1:}} To what extent do English-centric LLMs transfer mathematical reasoning across languages? (Section~\ref{sec:observation_study})
    \item \textit{\textbf{RQ2:}} What factors are associated with stronger or weaker cross-lingual transfer under controlled post-training settings? (Section~\ref{sec:interventional_study})
    \item \textit{\textbf{RQ3:}} How can cross-lingual transfer be improved effectively? (Section~\ref{sec:parallel_scaling_law})
\end{itemize}

\section{Observational Study}
\label{sec:observation_study}

To answer \textit{\textbf{RQ1}}---\emph{To what extent do English-centric LLMs transfer mathematical reasoning across languages?}---we begin with an observational study of open-source reasoning models. This analysis establishes whether cross-lingual transfer is present and examines how it varies with model initialization, target language, and training paradigm.

\subsection{Setup}

\paragraph{Models.}
We evaluate open-source reasoning models tuned with either SFT or RLVR, including the Simple-Zoo~\citep{zeng2025simplerl}, s1~\citep{muennighoff2025s1}, OpenThinker~\citep{guha2026openthoughts}, Open-Reasoner-Zero~\citep{hu2026open}, and DeepSeek-R1-Distill~\citep{guo2025deepseek} series. 
% Details are provided in Appendix~\ref{appendix:an_overview_of_open-source_LRMs}.

\paragraph{Benchmarks.}
We evaluate on multilingual mathematical reasoning benchmarks, including the multilingual versions of MATH500~\citep{hendrycks2021measuring}, AIME2024, and AIME2025 from XReasoning~\citep{qi2025models}. We also include multilingual GPQA-Diamond~\citep{rein2023gpqa} from BenchMAX~\citep{huang2025benchmax} as an out-of-domain reasoning benchmark. The evaluation includes \emph{English(En), Spanish (es), Russian (ru), German (de), French (fr), Bengali (bn), Swahili (sw), Thai (th), Japanese (ja), Chinese (zh), and Telugu (te)}---yielding eleven evaluation languages in total. 
More details are given in Appendix~\ref{appendix:multilingual_reasoning_benchmarks}.

\begin{figure*}[!h]
\centering
\includegraphics[width=0.95\linewidth]{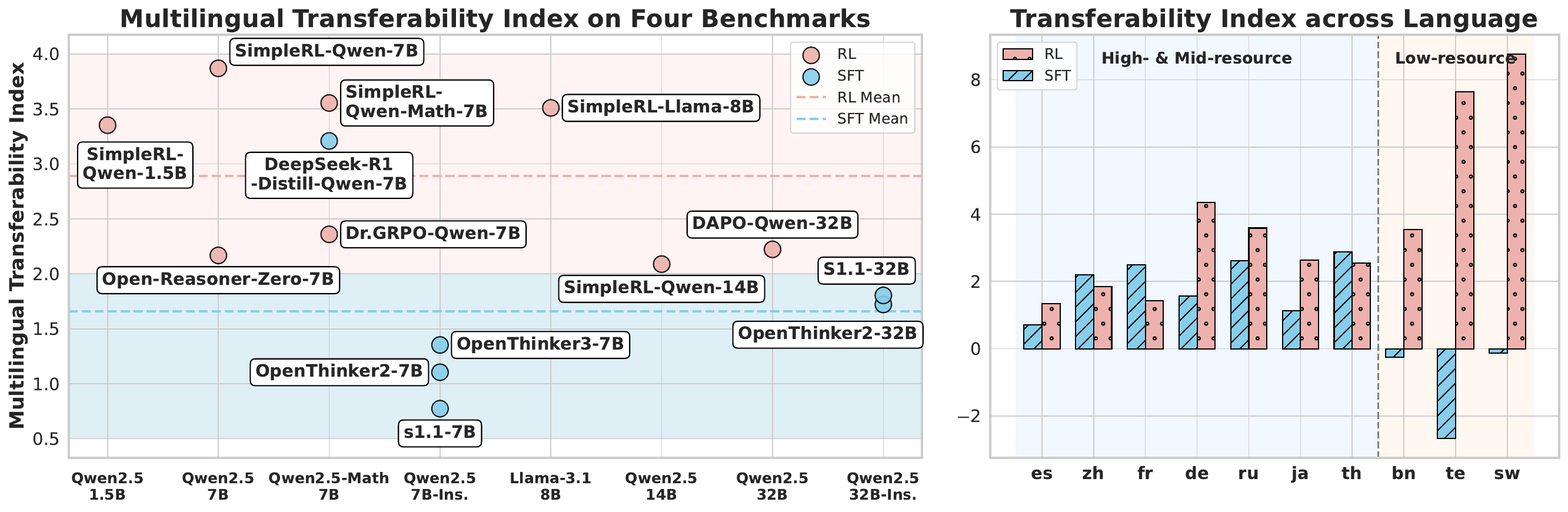}
\caption{\textbf{Cross-lingual transfer across open-source reasoning models.} The \emph{left} panel reports average MTI across four benchmarks and eleven languages. The \emph{right} panel shows average transfer behavior on multilingual MATH500, comparing SFT-tuned and RL-tuned models.}
\vspace{-4mm}
\label{fig:open_source_model_MTI}
\end{figure*}

\paragraph{Language-controlled prompting.}
Following prior work~\citep{luo2025mmath,wang2026polymath, qi2025models}, we concatenate the language control instructions after the input prompts to make the model generate responses using the same language as the query.
This setup reflects multilingual application scenarios where both the answer and the reasoning trace should be understandable in the user's language~\citep{yong2025crosslingual, wang2025language,zhang2026think}. The full prompts are provided in Appendix~\ref{prompt:multilingual_reasoning_instruction} and~\ref{appendix:prompt_hacking}.

\paragraph{Metrics.}
We report reasoning accuracy (\textbf{Acc}) as the main task metric, and off-target rate (\textbf{Off-target}) to measure how often a model fails to respond in the requested language, using the LangDetect library. In all evaluation settings, correctness is determined from the final answer only. Intermediate reasoning traces are not scored. Therefore, a model does not receive higher accuracy simply for producing its reasoning trace in the target language.

To quantify relative transfer, we also use the \emph{Multilingual Transferability Index} (MTI), following prior work~\cite{huan2025does,huang2026beyond}. MTI is intended to measure how much of the gain obtained in the training language transfers to unseen languages. Since base difficulty varies across languages and benchmarks, absolute accuracy alone is not sufficient for this purpose.

Let $Acc^{\text{trained}}_{b,l}$ and $Acc^{\text{base}}_{b,l}$ denote the accuracy of a trained model and its corresponding base model on benchmark $b$ and language $l$. We define the relative gain on language $l$ as
\begin{equation}
\label{eq:relative_gain}
\Delta R_{b,l} = \frac{Acc^{\text{trained}}_{b,l} - Acc^{\text{base}}_{b,l}}{Acc^{\text{base}}_{b,l}}.
\end{equation}
For a training language set $\mathcal{L}_{\text{train}}$, the average relative gain is
\begin{equation}
\Delta R_{b,\mathcal{L}_{\text{train}}} = \frac{1}{|\mathcal{L}_{\text{train}}|} \sum_{l \in \mathcal{L}_{\text{train}}} \Delta R_{b,l}.
\end{equation}
We define the MTI for an unseen language $l_{\text{unseen}}$ as the average benchmark-level relative transfer ratio:
\begin{equation}
\text{MTI}_{l_{\text{unseen}}}
=
\frac{1}{|B|}
\sum_{b \in B}
\frac{\Delta R_{b,l_{\text{unseen}}}}
{\Delta R_{b,\mathcal{L}_{\text{train}}}}.
\end{equation}

An MTI of 1 indicates that the relative gain on an unseen language matches the average gain on the training language(s); values below 1 indicate weaker relative transfer, and values above 1 indicate stronger relative transfer. 
MTI measures relative transfer efficiency rather than absolute multilingual reasoning ability: it captures how much of the training-language gain is reflected in unseen languages. 
Thus, a high MTI can still correspond to low absolute accuracy, especially from a low base accuracy.
For this reason, we use MTI only as an auxiliary diagnostic and interpret it together with absolute accuracy, non-English average accuracy, low-resource accuracy, and off-target rate throughout the paper, detailed in Appendix~\ref{appendix:absolute_transfer_metrics}.

% Because MTI is based on relative gain, it can be unstable when base accuracy is very low or when the gain on the training language is close to zero. 
% As a robustness check, we also repeated the main analyses with alternative transfer metrics, including normalized gain and raw gain ratio, and further examined the sensitivity of MTI to low-base-accuracy cases. The numerical values differ, but the main qualitative conclusions remain unchanged, in particular the large gain from monolingual to bilingual training (Appendix~\ref{appendix:mti_robustness}).
As a robustness check in Appendix~\ref{appendix:mti_robustness}, we also repeated the main analyses with alternative transfer metrics, including normalized gain and raw gain ratio, and examined the sensitivity of MTI to low-base-accuracy cases. The numerical values differ, but the main qualitative conclusions remain unchanged.

\subsection{Main Results}

The main result of this section is that \emph{cross-lingual transfer of mathematical reasoning is highly variable}. Figure~\ref{fig:open_source_model_MTI} shows that strong English reasoning does not reliably yield equally strong multilingual performance, and transfer varies across model initialization, training paradigm, and target language.

\paragraph{Model choice matters.}
Transfer varies substantially across models. Even under matched training settings, multilingual performance differs markedly across languages. For example, SimpleRL-Qwen-7B shows slightly higher transferability than SimpleRL-Qwen-Math-7B, despite using the same training data, training paradigm, and hyperparameters.

\paragraph{English performance is not enough.}
High English performance and strong multilingual transfer are related, but not identical. Some models that perform well in English do not maintain equally strong multilingual performance, while others with weaker English starting points show larger relative gains after post-training. English evaluation alone is therefore insufficient for characterizing multilingual transfer.

\paragraph{Training paradigm matters.}
In the open-source model comparison, RL-tuned models generally show stronger multilingual performance than SFT-tuned models, especially in lower-resource languages such as \emph{bn}, \emph{sw}, and \emph{te}. In these languages, SFT-tuned models often show weak or negative transfer, whereas RL-tuned models more often yield positive gains. Because the compared models also differ in initialization, training data, and implementation details, we treat this as an observational pattern rather than a controlled comparison of SFT and RL. A matched SFT-vs.-GRPO comparison on Qwen2.5-7B-Instruct is provided in Appendix~\ref{appendix:sft_vs_grpo}.

\paragraph{Target language matters.}
Transfer is not uniform across languages. Higher-resource languages are generally more stable, whereas lower-resource languages show larger relative variation across models and training paradigms. This suggests that cross-lingual transfer depends not only on English-side training, but also on properties of the target language.

% \subsection{Takeaway}

% The observational study establishes the motivation for the rest of the paper. English-centric reasoning models do exhibit non-trivial cross-lingual transfer, but that transfer is inconsistent: it varies substantially across models, languages, and post-training paradigms, and cannot be inferred from English performance alone. In Section~\ref{sec:interventional_study}, we move to controlled post-training settings to test which factors are systematically associated with this variation.

\section{Controlled Comparative Study}
\label{sec:interventional_study}

The observational study shows that cross-lingual transfer is highly variable, but open-source model comparisons cannot isolate the sources of this variation because models differ in initialization, data, optimization, and model family. To answer \textbf{RQ2}, we conduct controlled studies of initialization type, model family, and model size under matched GRPO-based post-training.

\subsection{Setup}

\paragraph{Training data.}
We use a compact training set of 1{,}000 examples sampled from the MATH training split, following prior work on small but carefully designed reasoning supervision sets~\citep{ye2025limo, muennighoff2025s1}. All controlled studies in this section use the same training set unless otherwise noted. Details are provided in Appendix~\ref{appendix:training_dataset}.

\paragraph{Post-training algorithm.}
% We use Group Relative Policy Optimization (GRPO)~\citep{shao2024deepseekmath} as our RPT algorithm. In our setup, the policy is optimized with a composite reward that combines reasoning accuracy, response format, and language consistency:
% \begin{equation}
%     R=\lambda_1R_{\text{acc}}+\lambda_2R_{\text{format}}+\lambda_3R_{\text{lang}}.
% \end{equation}
% Here, $R_{\text{acc}}$ rewards correctness, $R_{\text{format}}$ encourages well-formed reasoning traces, and $R_{\text{lang}}$ encourages responses in the target language. Hyperparameter details are given in Appendix~\ref{appendix:hyperparameters} and~\ref{sec:impact_of_reward_hyperpameters}.

% Because the reward includes language consistency and formatting terms in addition to task accuracy, post-training gains should not be interpreted as abstract reasoning gains alone. In this section, we use the same objective across settings so that comparisons remain controlled.

We use Group Relative Policy Optimization (GRPO)~\citep{shao2024deepseekmath} as our RPT algorithm. Following prior work~\cite{rastogi2025magistral,zhang2026think}, the policy is optimized with a composite reward that combines reasoning accuracy, response format, and language consistency:
\begin{equation}
    R=\lambda_1R_{\text{acc}}+\lambda_2R_{\text{format}}+\lambda_3R_{\text{lang}}.
\end{equation}
Here, $R_{\text{acc}}$ rewards correctness, $R_{\text{format}}$ encourages well-formed reasoning traces, and $R_{\text{lang}}$ encourages responses in the target language. Hyperparameter details are given in Appendix~\ref{appendix:hyperparameters} and~\ref{sec:impact_of_reward_hyperpameters}.

Because the objective includes formatting and language-consistency terms, gains should not be interpreted as pure reasoning gains alone; we keep the objective fixed across settings for controlled comparison.

\paragraph{Controlled factors.}
% We study four factors: \textit{training paradigm}, \textit{initial model type}, \textit{model family}, and \textit{model size}. In each case, all training data and optimization settings are fixed.

We study three factors: \textit{initial model type}, \textit{model family}, and \textit{model size}. In each case, all training data and optimization settings are fixed.

% \subsection{Training Paradigm}

% We first examine whether the observed transfer patterns also depend on the post-training paradigm. Using Qwen2.5-7B-Instruct as the common initialization, we compare SFT and GRPO under English-only training. For SFT, the supervision data are constructed from responses sampled from Qwen3-32B on the same English training set, whereas GRPO directly optimizes the target model with reward-based post-training. As shown in Table~\ref{tab:experiment_6_sft_rl}, both methods improve over the initial model, but GRPO yields larger gains in both multilingual accuracy and MTI. In particular, GRPO improves average accuracy from 22.45 to 23.51 and achieves a substantially higher MTI than SFT (1.23 vs.\ 0.52), while also slightly reducing the off-target rate. Under this setup, RL-based post-training transfers more effectively across languages than SFT.

% \input{tabs/experiment_6_sft_rl}

\subsection{Initial Model Type}

We first study the effect of initial model type using three Qwen2.5-7B variants: a general base model, an instruction-tuned model, and a math-specialized model. Table~\ref{tab:experiment_2_control_study_initial_model_simplify} shows a trade-off between final multilingual performance and relative transfer efficiency. The instruction model achieves the highest multilingual accuracy after English-only post-training and the lowest off-target rate, but the base and math-specialized models obtain higher MTI. This suggests that stronger final multilingual performance and stronger relative transfer efficiency are not always aligned under our setup. Notably, this observation concerns relative transfer efficiency rather than final multilingual performance, which remains highest for the instruction-tuned model under our setup.

\input{tabs/experiment_2_control_study_initial_model_simplify}
\subsection{Model Family}

We next compare Qwen2.5-7B-Instruct~\citep{qwen2025qwen25technicalreport} and Llama3.1-8B-Instruct~\citep{grattafiori2024llama3herdmodels} under the same English-only GRPO setup to test whether the pattern above persists across model families.

\begin{figure*}[!h]
\centering
\includegraphics[width=1.0\linewidth]{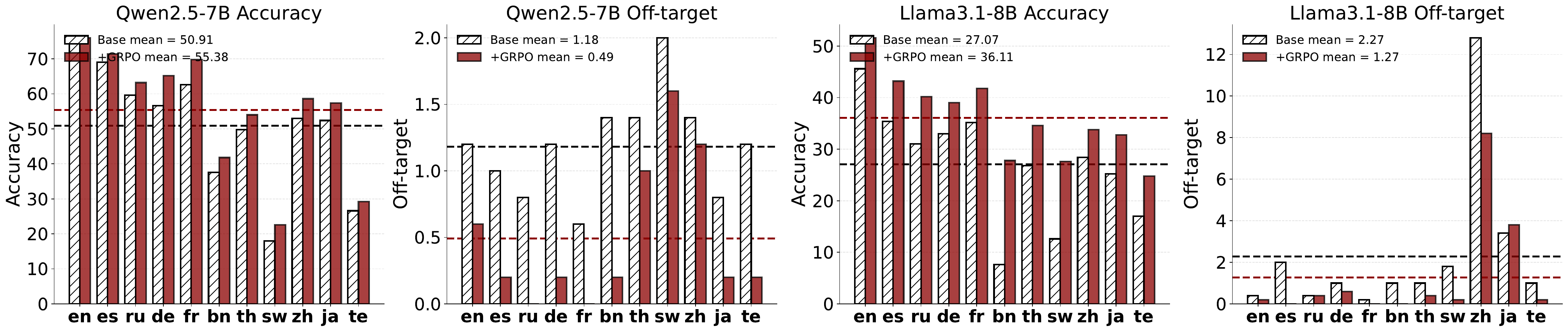}
\caption{\textbf{Effect of model family under controlled post-training.} Multilingual reasoning performance on MATH500. \emph{Base} denotes the initial model, and \emph{+GRPO} the model after English-only GRPO.}
\vspace{-4mm}
\label{fig:controlled_exp_different_initial_model_qwen_vs_llama}
\end{figure*}

Figure~\ref{fig:controlled_exp_different_initial_model_qwen_vs_llama} shows that English-only GRPO improves performance in both English and unseen languages for both families. However, the relative gains differ substantially: although Llama3.1-8B-Instruct starts from weaker English performance than Qwen2.5-7B-Instruct, it exhibits larger relative gains across languages. This suggests that cross-lingual transfer is not simply determined by English performance alone. At the same time, this comparison does not isolate a pure family effect, since differences in tokenization, pretraining data, and architecture remain entangled. Notably, this comparison concerns relative gains rather than final multilingual accuracy, and should be interpreted accordingly.

\subsection{Model Size}

We finally compare Qwen2.5-1.5B-Instruct and Qwen2.5-7B-Instruct under the same English-only GRPO setup to study how the observed pattern changes with model scale.

\begin{figure*}[!h]
\centering
\includegraphics[width=1.0\linewidth]{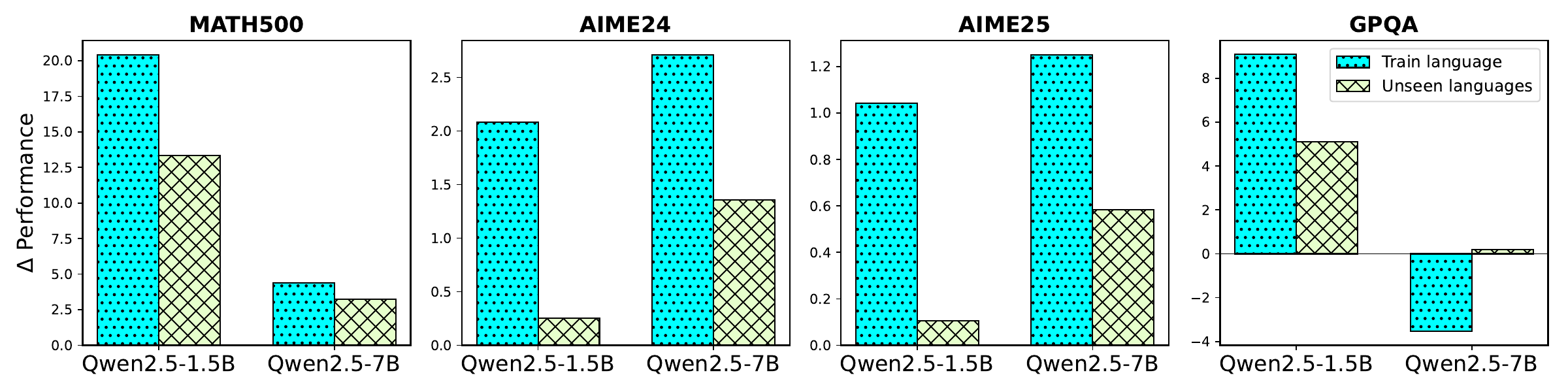}
% \caption{\textbf{Effect of model size under controlled post-training.} $\Delta$Performance denotes the average accuracy gain over the initial model, averaged across the training language and unseen languages.}
\caption{\textbf{Effect of model size under controlled English-only post-training.}
$\Delta$Performance denotes the average accuracy gain over the initial model, averaged across English and unseen languages.}
\label{fig:controlled_exp_scaling_up_model_size}
\vspace{-4mm}
\end{figure*}

Figure~\ref{fig:controlled_exp_scaling_up_model_size} shows that the effect of model scale is benchmark-dependent. On multilingual MATH500, the 1.5B model gains more than the 7B model, suggesting greater headroom on the in-domain task. On multilingual AIME24/25, however, the 7B model attains stronger final performance on the harder benchmarks despite smaller gains on MATH500. On multilingual GPQA-Diamond, the smaller model again shows larger improvements, while the 7B model improves only marginally and can even degrade in English. One possible explanation is that the small math-only training set over-specializes the 7B model toward mathematical reasoning, which can hurt out-of-domain scientific QA. Overall, the observed transfer pattern depends on the interaction between initial model strength and benchmark difficulty, rather than on model size alone.

% \subsection{Takeaway}

% The controlled studies sharpen the picture from Section~\ref{sec:observation_study}. Cross-lingual transfer in English-centric post-training is not well explained by English performance alone. Instead, it is systematically associated with initialization and alignment choices under controlled settings. Initial model type changes the balance between in-language usability and relative transfer efficiency, model family affects the magnitude of transfer, and model size interacts with benchmark difficulty. Across these comparisons, stronger English-centric initialization does not automatically imply stronger relative cross-lingual transfer.

% These findings motivate the final question of the paper: if English-only post-training yields unstable transfer, can it be improved directly? In the next section, we show that adding parallel multilingual supervision provides a simple and effective intervention.

\begin{figure*}[!h]
\centering
\includegraphics[width=1.0\linewidth]{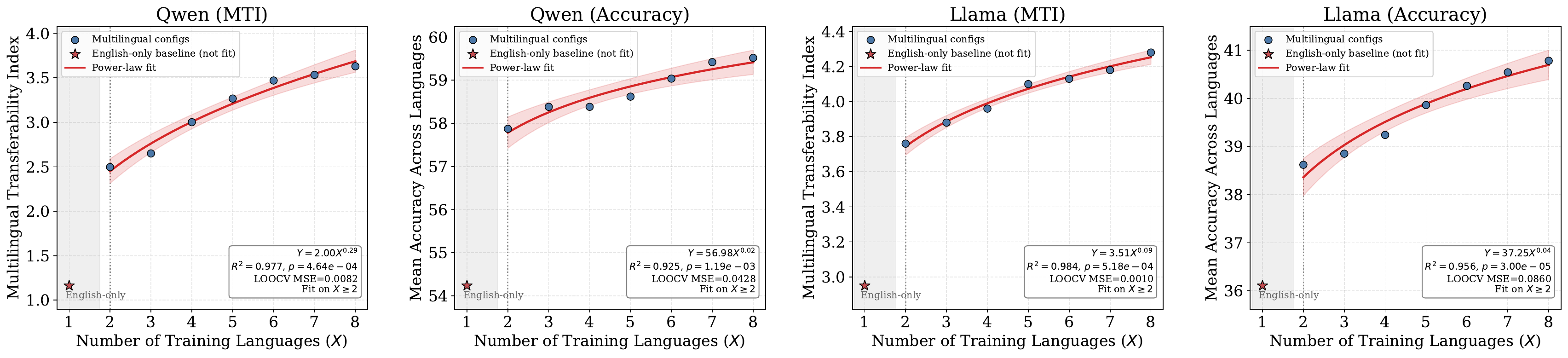}
\caption{
\textbf{Transfer improves with parallel multilingual supervision.}
Each panel shows performance as a function of the number of training languages. The red star denotes the English-only baseline ($X=1$), shown for reference but excluded from the fit. Blue points denote multilingual configurations ($X\geq2$). The solid red curve shows a power-law-like descriptive fit over $X\geq2$, with the shaded region indicating the $95\%$ confidence interval. The fit summarizes the bilingual-to-multilingual trend; Appendix~\ref{appendix:scaling_fit_analysis} provides the marginal-gain analysis.
}
\vspace{-4mm}
\label{fig:parallel_exp_scaling_law}
\end{figure*}

\section{Parallel Training Study}
\label{sec:parallel_scaling_law}

The controlled comparative results suggest that cross-lingual transfer in English-centric post-training is unstable and sensitive to initialization and alignment choices. We therefore ask \textit{\textbf{RQ3}}---\emph{How can cross-lingual transfer be improved effectively?} In this section, we study a simple intervention: adding multilingual parallel supervision during post-training.

Our main finding is that the monolingual-to-bilingual transition yields the largest observed single-step improvement. Additional languages provide smaller but positive gains.

\subsection{Setup}

% We use Qwen2.5-7B-Instruct as the main initial model and post-train it with GRPO on English together with increasing numbers of parallel languages. The starting point is the same 1{,}000 English examples used in Section~\ref{sec:interventional_study}. For multilingual supervision, we use parallel problem sets drawn from the multilingual MATH dataset described in Section~\ref{appendix:multilingual_reasoning_benchmarks}, covering seven additional languages: \emph{es, ru, de, fr, bn, th, zh}. Following prior work, we further perform an LLM-based validation with Qwen3-32B to screen for translation errors in the multilingual dataset; details and summary statistics are provided in Appendix~\ref{appendix:multilingual_reasoning_benchmarks}. By varying the number of added languages from one to seven, we obtain a sequence of models that differ only in the amount of multilingual supervision. The full language configurations are given in Table~\ref{tab:appendix9_scaling_parallel_law_languages}.
% We evaluate these models on multilingual MATH500 across eleven languages, using absolute accuracy as the primary metric and MTI only as an auxiliary measure of relative transfer efficiency. Unless otherwise stated, the analyses in this section focus on multilingual MATH500, which provides the clearest view of the monolingual-to-bilingual transition under our setup.

We use Qwen2.5-7B-Instruct as the main initial model and post-train it with GRPO on the same 1{,}000 English MATH examples used in Section~\ref{sec:interventional_study}, together with increasing numbers of parallel languages. The parallel problem sets are drawn from the multilingual MATH dataset described in Appendix~\ref{appendix:multilingual_reasoning_benchmarks}, covering seven additional languages: \emph{es, ru, de, fr, bn, th, zh}. We validate the multilingual data with Qwen3-32B to screen for translation errors, with details and summary statistics reported in Appendix~\ref{appendix:multilingual_reasoning_benchmarks}. By varying the number of added languages from one to seven, we obtain a sequence of models that differ only in the amount of multilingual supervision. The full language configurations are given in Table~\ref{tab:appendix9_scaling_parallel_law_languages}.

We evaluate these models on multilingual MATH500 across eleven languages, using absolute accuracy as the primary metric and MTI as an auxiliary measure of relative transfer efficiency. Unless otherwise stated, this section focuses on multilingual MATH500, which provides the clearest view of the monolingual-to-bilingual transition in our setup.

\paragraph{The Power of Bilingual Training}

The most direct pattern in Figure~\ref{fig:parallel_exp_scaling_law} is that the largest observed single-step gain occurs when moving from English-only to bilingual parallel training. Using Qwen2.5-7B-Instruct as the initial model, average accuracy increases from 54.24 to 57.87, while MTI increases from 1.16 to 2.50. Importantly, this improvement is not limited to MTI. As shown in Appendix~\ref{appendix:absolute_transfer_metrics}, the same transition improves non-English average accuracy from 51.74 to 55.82, reduces the English--non-English gap from 27.46 to 22.58, and improves low-resource average accuracy from 29.53 to 36.27.

We refer to this observed pattern as the \textit{First-Parallel Leap}. This term denotes an empirical marginal-gain pattern in our setting: the first added parallel language accounts for the largest single-step improvement, while additional languages continue to provide smaller but positive gains. Thus, moving beyond English-only supervision is especially important, while multilingual benefits do not saturate after one added language.

\paragraph{Beyond Bilingual: Positive but Diminishing Returns}

Beyond the bilingual setting, performance continues to improve as more parallel languages are added, but the gains are smaller. Figure~\ref{fig:parallel_exp_scaling_law} shows a smooth sub-linear trend in both MTI and average accuracy. To summarize this pattern, we fit a power-law-like function
\begin{equation}
f(X)=\alpha X^{\beta},
\end{equation}
where $X$ is the number of training languages. We fit only configurations with $X \geq 2$ to separate the monolingual and multilingual regimes. The resulting fits are:
\begin{equation}
\begin{aligned}
\textit{Accuracy:}\quad f(X) &= 56.98X^{0.02},\\
\textit{Transferability:}\quad f(X) &= 2.00X^{0.29}.
\end{aligned}
\end{equation}

We use this \textit{Parallel Scaling Trend} as a descriptive summary of the bilingual-to-multilingual regime, rather than as a universal scaling law. The sub-linear exponents indicate diminishing returns as more languages are added. To characterize the monolingual-to-bilingual transition, we rely primarily on observed marginal gains: the largest single-step improvement occurs when moving from English-only to bilingual parallel training, while additional languages continue to yield smaller but positive gains. Appendix~\ref{appendix:scaling_fit_analysis} compares alternative functional forms and reports influence diagnostics, supporting the use of the power-law-like curve as a compact descriptive fit for the multilingual regime.

% We refer to this descriptive pattern as the \textit{Parallel Scaling Trend}. We use it to summarize the bilingual-to-multilingual regime in our experiments, not to claim a universal scaling law. The sub-linear exponents indicate diminishing returns as more languages are added.

% \paragraph{The Monolingual Extrapolation Gap}

% The fitted multilingual trend also provides a comparison point for English-only training. When the bilingual-to-multilingual fit is extrapolated back to $X=1$, the observed monolingual baseline falls below the extrapolated value in both MTI and accuracy. For example, the extrapolated MTI at $X=1$ is about 2.00, while the observed English-only value is 1.16. A similar gap appears for accuracy.

% We refer to this discrepancy as the \textit{Monolingual Extrapolation Gap}. The interpretation is descriptive rather than mechanistic: English-only post-training behaves differently from the multilingual regime summarized by the fitted trend.

\subsection{Additional Analysis and Discussion}

\paragraph{Fixed training budget.}

A possible alternative explanation is that parallel training helps simply because it uses more data. To test this, we fix the total number of training examples and vary only the composition. Specifically, we compare scaling English-only data (e.g., $2N$ English examples) with replacing part of it by parallel multilingual data (e.g., $N$ English + $N$ Russian). Here, $2N$ and $3N$ denote matched budgets of unique training examples rather than repeated epochs over the same $N$ examples. Table~\ref{tab:experiment_4_fixed_budget} shows that multilingual mixtures yield stronger cross-lingual transfer than monolingual scaling alone.

\input{tabs/experiment_4_fixed_budget}

\paragraph{Parallel vs.\ non-parallel data.}

\begin{figure*}[!h]
\centering
\includegraphics[width=1.0\linewidth]{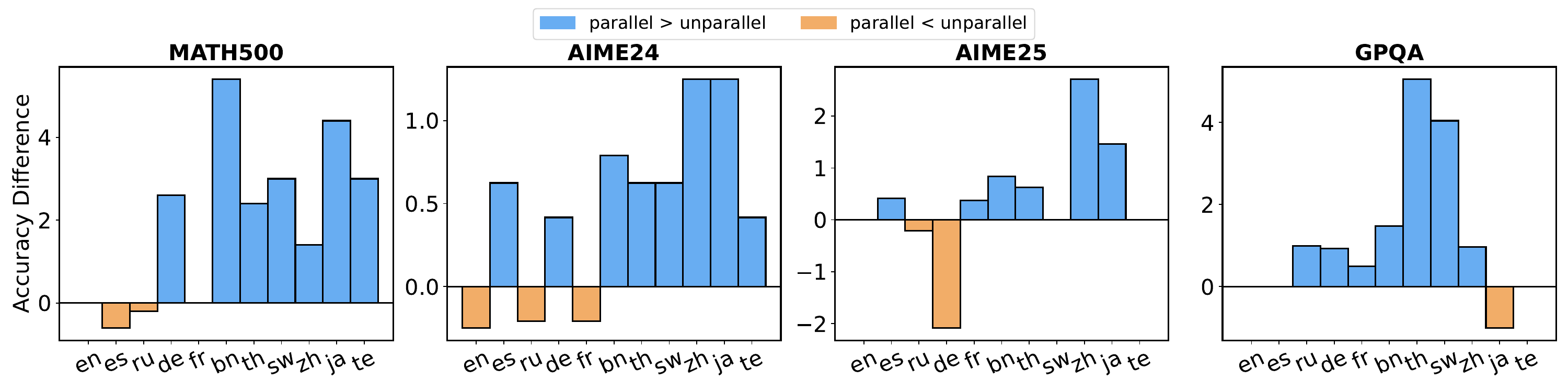}
\caption{\textbf{Parallel vs.\ non-parallel multilingual supervision.} Accuracy differences under parallel and non-parallel data training based on Qwen2.5-7B-Instruct.}
\vspace{-4mm}
\label{fig:parallel_vs_unparallel}
\end{figure*}

We also compare parallel bilingual supervision against non-parallel multilingual data.
This comparison separates multilingual exposure from cross-lingual alignment. In the parallel setting, the model sees aligned versions of the same mathematical problems across languages, whereas in the non-parallel setting, it sees different problems in different languages and therefore has greater problem diversity. An exposure- or diversity-based explanation would predict comparable gains from non-parallel multilingual supervision. Figure~\ref{fig:parallel_vs_unparallel} shows that parallel data yields stronger gains, suggesting that same-problem cross-lingual alignment is an important part of the bilingual improvement. Additional group-level statistics are provided in Appendix~\ref{appendix:parallel_vs_nonparallel}.

\paragraph{Does the chosen language matter?}

\begin{figure*}[!h]
\centering
\includegraphics[width=1.0\linewidth]{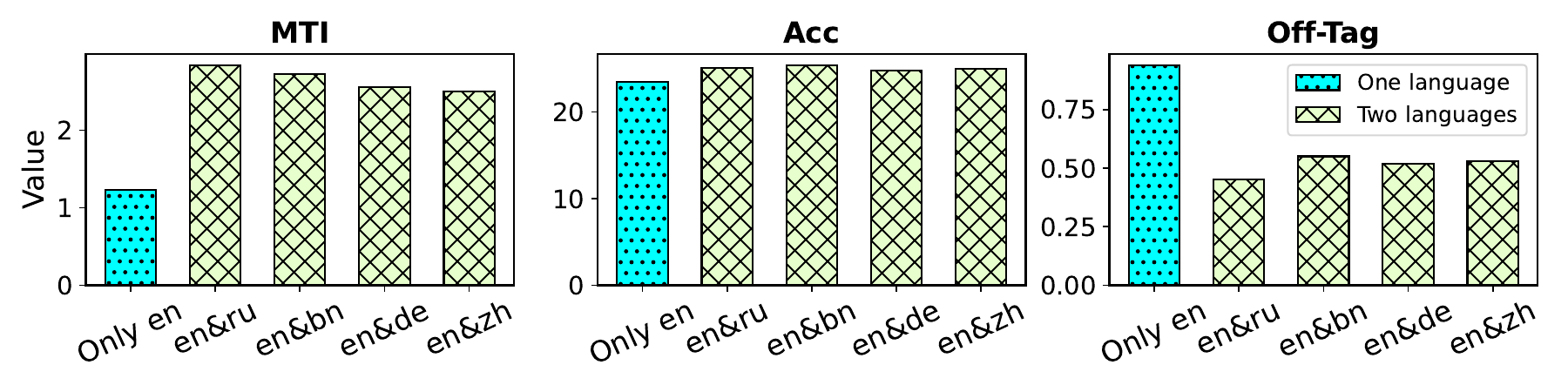}
\caption{\textbf{The impact of the selected parallel language.} Multilingual reasoning performance across different single-language parallel training settings based on Qwen2.5-7B-Instruct.}
\vspace{-4mm}
\label{fig:parallel_exp_selected_langs}
\end{figure*}

A practical question is whether the gain depends strongly on which bilingual language is added. Figure~\ref{fig:parallel_exp_selected_langs} shows that adding one parallel language consistently improves transfer, with modest differences across Russian, Bengali, German, and Chinese. Thus, the bilingual gain is robust to the choice of added language, although some language-specific effects remain.

\paragraph{Scaling the amount of training data}

\begin{figure}[!h]
\centering
\includegraphics[width=1.0\linewidth]{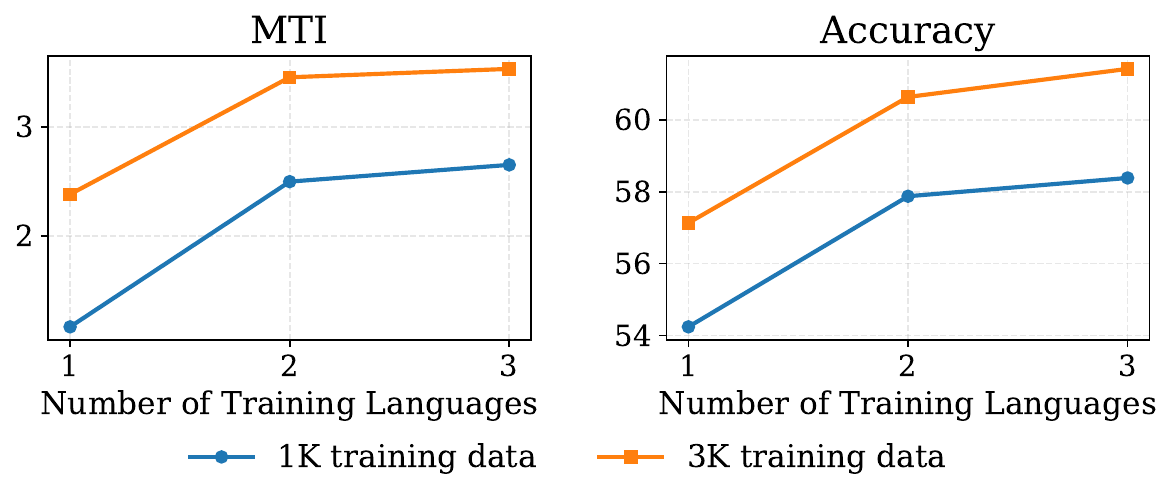}
\caption{Scaling training data based on Qwen2.5-7B-Instruct.}
\label{fig:scaling_training_data}
\vspace{-4mm}
\end{figure}

We also study whether the benefit of parallel training persists as the amount of training data increases. Figure~\ref{fig:scaling_training_data} compares the 1K and 3K training data for each language under English-only, bilingual, and trilingual training settings. Increasing the training data improves both MTI and average accuracy, while the relative pattern remains stable: the largest gain comes from moving from monolingual to bilingual training, and adding a second parallel language yields a smaller additional improvement.

\paragraph{Prompt ablation.}
A possible concern is that the bilingual gains in the main setting are driven by explicit language-control prompting. To test this, we compare two evaluation protocols: \emph{language-controlled prompting}, used in the main experiments, and \emph{no language control}, where all explicit target-language instructions are removed. We evaluate two Qwen2.5-7B-Instruct-based models: an \emph{English-only} GRPO model and an \emph{English+Ru} parallel-trained GRPO model. Table~\ref{tab:measurement_disentanglement_prompt} shows that the bilingual gains are not reducible to test-time prompt control alone. Without language-control prompts, the English-only model exhibits severe language drift, whereas the parallel-trained model maintains near-zero off-target behavior and slightly higher accuracy. Under language-controlled evaluation, the parallel-trained model also remains stronger in accuracy.
\input{tabs/experiment_7_prompt_ablation}

\paragraph{Extension across model families and source languages.}
We repeat the analysis on Llama3.1-8B-Instruct and observe the same qualitative pattern in Figure~\ref{fig:parallel_exp_scaling_law}: the largest gain comes from moving beyond English-only training. We further test a Chinese-centric variant in Table~\ref{tab:chinese_parallel_scaling_main}. The same trend appears there as well: adding one parallel language yields the largest improvement, while adding another language provides a smaller additional gain. These results suggest that the bilingual gain is not specific to a single model family or source language.

\input{tabs/experiment_8_chinese_centric}

% \paragraph{Language reward ablation}

% A possible concern is that the gains from parallel training are partly driven by the language-consistency reward. To examine this, we vary the weight of $R_{\mathrm{lang}}$ under English+Russian parallel training. Table~\ref{tab:appendix_10_the_impact_of_hyperparameter_lambda_simplify} shows that the default setting performs best overall. Removing $R_{\mathrm{lang}}$ substantially increases the off-target rate and lowers MTI, while increasing its weight does not further improve the main results. This suggests that the gains from parallel training are not explained by over-weighting language adherence alone, although reward-side analysis cannot fully isolate pure reasoning transfer.

% \input{tabs/appendix_10_the_impact_of_hyperparameter_lambda_simplify}

% \paragraph{Language reward ablation.}
% A possible concern is that the gains from parallel training are partly driven by the language-consistency reward. To assess this, we vary the weight of $R_{\mathrm{lang}}$ under English+Russian parallel training. The default setting performs best overall: removing $R_{\mathrm{lang}}$ substantially increases the off-target rate, while increasing its weight does not improve the main results. This suggests that the bilingual gain is not explained by over-weighting language adherence alone, although reward-side analysis

\paragraph{Robustness to transfer metric choice.}
We further examine whether the observed monolingual-to-bilingual gain depends on the specific formulation of MTI. To this end, we repeat the main parallel-training analysis using two alternative transfer metrics, normalized-gain ratio and raw-gain ratio in Table~\ref{tab:metric_robustness_summary}, and additionally test the sensitivity of the results to low-base-accuracy cases by excluding unseen-language cases below different base-accuracy thresholds in Table~\ref{tab:mti_base_sensitivity}. While the absolute values differ across metrics and filters, the main qualitative result remains unchanged: the largest improvement still comes from moving from monolingual to bilingual training. This suggests that the bilingual gain is not specific to the original MTI definition (Appendix~\ref{appendix:mti_robustness}).

\section{Related Work}

\paragraph{Reasoning Transfer}
Recent large reasoning models (LRMs) have been driven by reinforcement-learning-based post-training. OpenAI’s O1~\citep{jaech2024openai} and DeepSeek-R1~\citep{guo2025deepseek} show that RL can substantially improve mathematical reasoning, especially when combined with verifiable or rule-based rewards~\citep{shao2024deepseekmath}. As these models improve, an important question is how well the acquired reasoning behavior transfers beyond the training setting. Prior work has mainly studied this problem across tasks~\citep{hu2026breaking}, domains~\citep{huan2025does,chu2025sft}, or modalities~\citep{liu2025x}, showing that RL-induced gains may not transfer uniformly beyond the training setting.
% \citet{hu2026breaking} show that RL improves structured reasoning but transfers less effectively to unstructured tasks. \citet{huan2025does,chu2025sft} compare RL and SFT in terms of broader transfer rather than narrow task-specific gains, and X-REASONER~\citep{liu2025x} studies transfer across domains and modalities. 
Our work is complementary to this line of research: instead of cross-task or cross-modal transfer, we focus on cross-lingual transfer of mathematical reasoning.

\paragraph{Cross-Lingual Transfer}
Cross-lingual transfer in English-centric LLMs has been widely studied in multilingual NLP, alignment, and reasoning settings. Prior work shows that English-trained reward models~\citep{wu2024reuse,hong2025cross}, preference alignment~\citep{yang2025language,yang2025implicit}, and lightweight multilingual adaptation~\citep{li2024x,chirkova2024zero} can often generalize beyond English. In multilingual reasoning, recent studies examine model merging for mathematical capability transfer~\citep{bandarkar2025layer}, cross-lingual test-time scaling~\citep{yong2025crosslingual}, response-language control~\citep{qi2025models}, language--reasoning disentanglement~\citep{zhao2026less}, and RL training on non-English data~\citep{huang2026beyond}. Parallel multilingual data has also been shown to improve multilingual capability in prompting and pretraining settings~\citep{mu2024revealing,qorib2025just}. Our work is complementary: we study cross-lingual transfer in reasoning-oriented post-training, compare same-problem parallel supervision with non-parallel multilingual exposure, and analyze how transfer changes as the number of parallel training languages increases.

\section{Conclusion}

We study cross-lingual mathematical reasoning transfer in English-centric LLMs under multilingual evaluation. Our results show that transfer exists but is highly variable, and that English reasoning performance alone does not reliably predict multilingual performance. Through controlled GRPO-based experiments, we find that stronger initialization can improve final multilingual accuracy without necessarily improving relative transfer efficiency. We further show that parallel multilingual supervision improves transfer in our setting, with the largest observed single-step gain from English-only to bilingual training. Overall, our findings highlight the need to evaluate multilingual reasoning directly using both absolute performance and relative transfer-efficiency metrics.

\section*{Acknowledgements}

We thank the anonymous reviewers for their insightful feedback and constructive comments. 
This work is supported by Beijing Natural Science Foundation L259016 and the Strategic Priority Research Program of Chinese Academy of Sciences under Grant XDA04080400.

\section*{Limitations}

Our study has several limitations. First, our controlled experiments remain small-scale compared with production-level reasoning post-training, even though we extend the training data from 1K to 3K examples and observe the same qualitative pattern. The results should therefore be interpreted as evidence under our controlled setup rather than as claims about large-scale post-training dynamics. In addition, our controlled experiments are primarily based on GRPO, so they should not be read as a general comparison of all RL-style post-training methods against SFT.
Second, our multilingual evaluation relies on translated benchmarks, which may introduce language-dependent artifacts such as wording shifts, notation mismatch, or uneven difficulty across languages. Evaluation is also conducted under language-controlled prompting, so the reported gains may reflect language compliance and response-format stability in addition to task-solving ability.
Finally, MTI is used only as an auxiliary diagnostic of relative transfer efficiency. It can be unstable in low-accuracy regimes and should not be interpreted as a measure of absolute multilingual reasoning ability. Similarly, the power-law-like trend in the parallel-training study is intended as a descriptive summary of the observed bilingual-to-multilingual regime, rather than as a theoretical scaling law. Overall, our controlled results establish empirical associations rather than a definitive causal mechanism.

\section*{Ethical Considerations}

This work studies the cross-lingual transfer of mathematical reasoning in English-centric LLMs. Our contributions are primarily empirical and methodological: we analyze transfer behavior under controlled multilingual evaluation and study training interventions that may improve multilingual reasoning performance. We use publicly available benchmarks that do not contain sensitive personal information to the best of our knowledge. At the same time, multilingual evaluation based on translated benchmarks may overstate or understate performance in some languages if translation artifacts, notation mismatch, or mixed-language mathematical responses are not carefully accounted for. Improvements in response-language control should therefore not be interpreted as complete evidence of robust multilingual reasoning in real-world settings.

\bibliography{reference}

\clearpage
\onecolumn
\appendix
\section*{Appendix}

\startcontents[sections]
\printcontents[sections]{l}{1}{\setcounter{tocdepth}{3}}

\clearpage
\twocolumn

\section{Reproducibility Statement}
Code and model weights will be released upon publication.
For evaluation, we follow prior works and report averaged results over 16 sampled generations per question on data-scarce benchmarks.
All evaluations are conducted with temperature set to 0.6 and top-p to 0.95, with the random seed fixed to ensure deterministic outputs across runs.
Note that minor variations in inference results may still occur due to differences in hardware or the version of the inference framework.

\section{The Usage of AI Assistants}
We declare that the AI Assistants (ChatGPT and Gemini) were only used to refine the fluency of certain sentences during the writing of this paper. Every sentence polished with the LLM was carefully reviewed and approved by the authors. The LLM was not used for any other part of this research.

\section{Evaluation Details and Setup}
\subsection{Multilingual Reasoning Benchmarks}
\label{appendix:multilingual_reasoning_benchmarks}
We use the multilingual version of these four reasoning benchmarks provided in~\cite{qi2025models}, which use GPT-4o~\citep{jaech2024openai} to translate all questions into the other ten languages \emph{Spanish (es), Russian (ru), German (de), French (fr), Bengali (bn), Swahili (sw), Thai (th), Japanese (ja), Chinese (zh), and Telugu (te)}, resulting in a total of eleven languages for evaluation.
\paragraph{MATH500}
The MATH500~\citep{hendrycks2021measuring} benchmark assesses the mathematical reasoning and problem-solving abilities of language models, addressing the need for more challenging evaluations as their general capabilities advance. It consists of 500 problems across five core mathematical domains: algebra, combinatorics, geometry, number theory, and precalculus. Each problem is designed to test multi-step reasoning and complex problem-solving skills, going beyond simple calculations or factual recall.

\paragraph{AIME24\&25} 
The AIME24 and AIME25 datasets contain problems from the American Invitational Mathematics Examination (AIME) for 2024 and 2025, respectively. AIME is a prestigious high school mathematics competition renowned for its challenging problems, consisting of 30 questions.

\paragraph{GPQA-Diamond}
GPQA-Diamond~\citep{rein2023gpqa} consists of 198 multiple-choice questions across biology, chemistry, and physics, with difficulty levels ranging from challenging undergraduate to postgraduate. It is the highest quality subset, which includes only questions where both experts answer correctly and the majority of non-experts answer incorrectly.

\paragraph{Translation Validation.}
To assess the quality of the translated multilingual benchmarks, we perform a full-coverage LLM-based validation following prior work~\cite{zhang2026think}. For each benchmark and each target language, we compare every translated question against its English source using Qwen3-32B as a judge. This validation is conducted per question rather than by sampling. The judge scores each translation along four dimensions: semantic fidelity, mathematical consistency, solvability preservation, and fluency/naturalness. Based on these scores, it assigns an overall label of \emph{Valid}, \emph{Minor issue}, or \emph{Major issue}, together with a short rationale and an error type when applicable (e.g., semantic shift, omission, mathematical inconsistency, or ambiguity). Across all checked translations, 99.2\% are labeled as \emph{Valid}, 0.7\% as \emph{Minor issue}, and 0.1\% as \emph{Major issue}. We use this procedure to screen for obvious translation errors and to verify that the translated benchmarks preserve the original mathematical problems at scale.

\subsection{Evaluation Metrics}
\label{appendix:evaluation_metrics}

We use three types of evaluation metrics: task accuracy, response-language consistency, and relative transfer efficiency.

\paragraph{Accuracy.}
Accuracy measures whether the final answer is correct. Following standard mathematical reasoning evaluation, correctness is determined from the final answer only; intermediate reasoning traces are not separately scored.

\paragraph{Language consistency.}
To measure whether a model responds in the requested language, we report the off-target rate. Following LCR-style diagnostics used in multilingual reasoning benchmarks such as PolyMath and MMATH, we use LangDetect to identify the language of each generated response. Let $\hat{l}_i$ denote the language detected from the $i$-th generated response, and let $l_i$ denote the requested target language. For a set of $N$ responses, the language consistency rate (LCR) is defined as:
\begin{equation}
\mathrm{LCR}
=
\frac{1}{N}
\sum_{i=1}^{N}
\mathbb{I}\left[\hat{l}_i = l_i\right],
\end{equation}
where $\mathbb{I}[\cdot]$ is the indicator function. We define the off-target rate as the complement of LCR:
\begin{equation}
\mathrm{Off\text{-}target}
=
1-\mathrm{LCR}.
\end{equation}
Thus, lower off-target rate indicates higher response-language consistency. In this work, we report aggregate response-level off-target rate and do not separately score the language consistency of intermediate reasoning traces and final answers.

\paragraph{Relative transfer efficiency.}
We use the Multilingual Transferability Index (MTI) as an auxiliary diagnostic of relative transfer efficiency. MTI measures how much of the training-language gain is reflected in unseen languages. Let $Acc^{\text{trained}}_{b,l}$ and $Acc^{\text{base}}_{b,l}$ denote the accuracy of a trained model and its corresponding base model on benchmark $b$ and language $l$. We define the relative gain on language $l$ as:
\begin{equation}
\Delta R_{b,l} =
\frac{Acc^{\text{trained}}_{b,l} - Acc^{\text{base}}_{b,l}}
{Acc^{\text{base}}_{b,l}}.
\end{equation}
For a training language set $\mathcal{L}_{\text{train}}$, the average relative gain is:
\begin{equation}
\Delta R_{b,\mathcal{L}_{\text{train}}}
=
\frac{1}{|\mathcal{L}_{\text{train}}|}
\sum_{l \in \mathcal{L}_{\text{train}}}
\Delta R_{b,l}.
\end{equation}
For an unseen language $l_{\text{unseen}}$, MTI is then defined as:
\begin{equation}
\text{MTI}_{l_{\text{unseen}}}
=
\frac{1}{|B|}
\sum_{b \in B}
\frac{\Delta R_{b,l_{\text{unseen}}}}
{\Delta R_{b,\mathcal{L}_{\text{train}}}}.
\end{equation}

MTI should be interpreted as a relative transfer-efficiency metric rather than a measure of absolute multilingual reasoning ability. We therefore report MTI together with absolute accuracy, non-English average accuracy, low-resource accuracy, and off-target rate.

\subsection{An Overview of Open-source LRMs}
\label{appendix:an_overview_of_open-source_LRMs}
Table~\ref{tab:appendix1_open_source_model} provides an overview of the various open-source LLMs evaluated in our observational study. These models, which include the DeepSeek-R1-Distill-Qwen-7B~\citep{guo2025deepseek}, OpenThinker series~\citep{guha2026openthoughts}, Simple-RL-Zoo series~\citep{zeng2025simplerl}, s1 series~\citep{muennighoff2025s1}, and DAPO-Qwen-32B~\citep{yu2026dapo}, range in size from 1.5B to 32B.

\input{tabs/appendix_1_open_source_model}

\section{Implementation Details}

\subsection{GRPO Algorithm}
GRPO is a simplified PPO-based algorithm that significantly reduces training costs by eliminating the need for a value model. 
It operates by sampling $G$ rollouts $\{o_1,...,o_G\}$ from the current policy for a given input, calculating their cumulative rewards $R=\{R_1,...,R_G\}$, and then using these rewards to estimate advantages $\hat{A}_{i,t}$ to guide policy updates. 
The optimization objective for GRPO is defined as follows:
\begin{equation}
\begin{aligned}
\mathcal{L}^{\text{clip}}_{i,t}(\theta)
&= \min \Big(
r_{i,t}(\theta)\hat{A}_{i,t}, \\
&\quad
\operatorname{clip}\big(
r_{i,t}(\theta),
1-\varepsilon,
1+\varepsilon
\big)\hat{A}_{i,t}
\Big), \\
r_{i,t}(\theta)
&=
\frac{
\pi_{\theta}(o_{i,t}\mid q,o_{i,<t})
}{
\pi_{\theta_{\text{old}}}(o_{i,t}\mid q,o_{i,<t})
}, \\
\hat{A}_{i,t}
&=
\frac{
R_i-\operatorname{mean}(R)
}{
\operatorname{std}(R)
}.
\end{aligned}
\end{equation}
The clipping term with ratio $\varepsilon$~\citep{schulman2015trust} keeps the new policy close to the old one, improving training stability.

\subsection{Training Dataset}
\label{appendix:training_dataset}

\paragraph{The Distribution of Parallel Questions}

\begin{figure*}[!h]
\centering
\begin{subfigure}{0.495\linewidth}
    \centering
    \includegraphics[width=\linewidth]{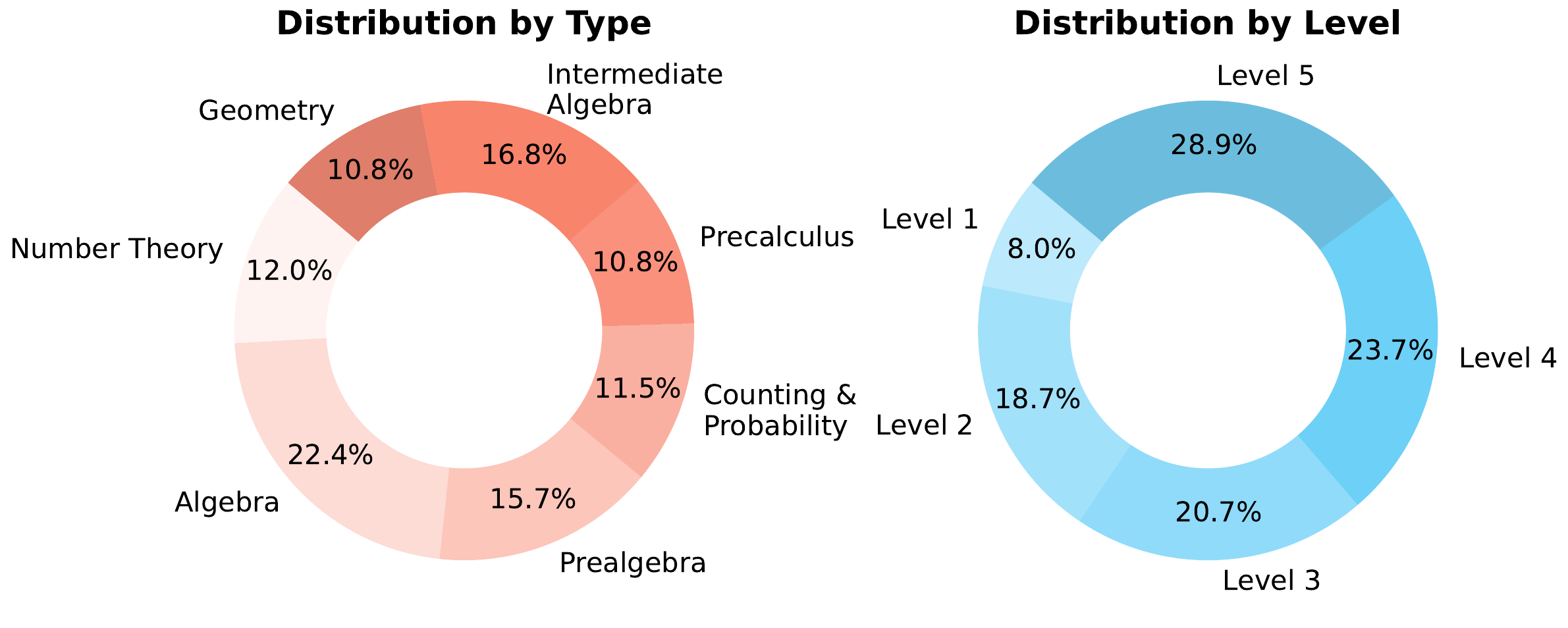}
    \caption{}
    \label{fig:stat_dataset_type_level_en}
\end{subfigure}
\begin{subfigure}{0.495\linewidth}
    \centering
    \includegraphics[width=\linewidth]{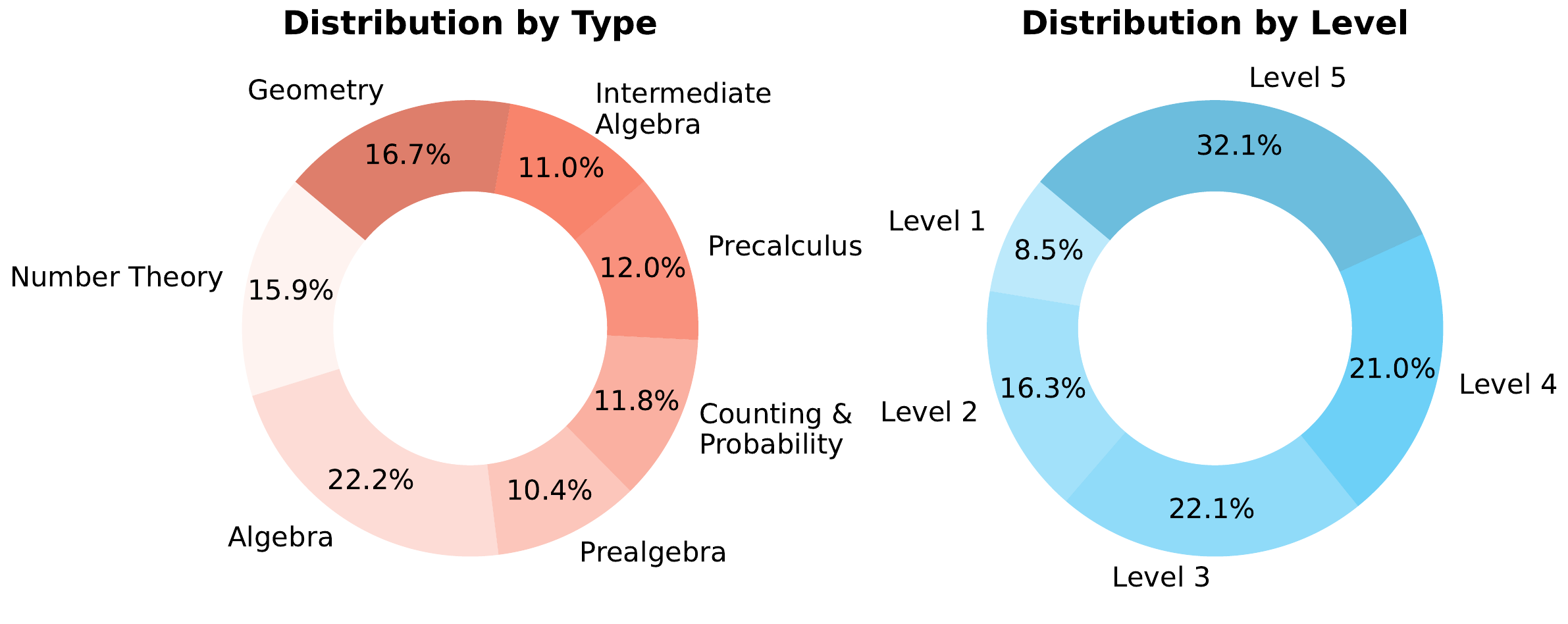}
    \caption{}
    \label{fig:stat_dataset_type_level_ru}
\end{subfigure}

\caption{Distribution of question difficulty. (a) The 1,000 English questions were utilized in the controlled comparative study and for parallel training. (b) The 1,000 Russian questions for non-parallel training, comprising a separate and non-overlapping set from the questions in (a).}
\label{fig:stat_dataset_type_level}
\end{figure*}

Figure~\ref{fig:stat_dataset_type_level_en} shows the type and level distributions of the 1,000 English training questions sampled from the MATH dataset~\citep{hendrycks2021measuring}. The type distribution is relatively balanced, and the number of questions increases steadily from Level 1 to Level 5.

\paragraph{The Distribution of non-parallel Questions}

Moreover, Figure~\ref{fig:stat_dataset_type_level_ru} presents the type and level distributions of 1,000 Russian questions used for a non-parallel training analysis experiment, which form a separate, non-overlapping set from the 1000 English questions. The distributions of both type and level closely match those of the English training questions. This indicates that, in the analysis comparing parallel and non-parallel training, the performance drop observed in non-parallel training is not due to distributional differences between the non-parallel and English datasets.

% \paragraph{Construct SFT Training dataset}

\subsection{Experimental Environments}
\label{appendix:experiments_environments}

Training and inference were conducted on Ubuntu 22.04 with 8 $\times$ NVIDIA A800 GPUs. We use VeRL~\citep{sheng2025hybridflow} for RL training, vLLM 0.8.5~\citep{kwon2023efficient} for inference, and Qwen's Math evaluation codebase~\citep{yang2024qwen2} following prior work~\citep{zeng2025simplerl, liu2025understanding}.

\subsection{Hyperparameters}
\label{appendix:hyperparameters}
\paragraph{RL Training} 
The maximum generation length was set to 4096 tokens, and the maximum prompt length to 1024 tokens, such that their sum matches the model’s maximum context length. The learning rate was fixed at $1\times10^{-6}$. Training was performed with a batch size of 128 questions. For each question, $G=16$ rollouts were sampled, using a sampling temperature of 1.0. $\lambda_1=0.8$, $\lambda_2=0.1$, and $\lambda_3=0.1$.

\paragraph{Inference}
In the evaluation setup, we used a temperature of 0.6, a top-$p$ value of 0.95, and a maximum generation length of 8912 tokens for all models in the 1.5B–14B series. For 32B models, we used the same temperature (0.6) and top-$p$ value (0.95), but set the maximum generation length to 32,768 tokens, except for DAPO-Qwen-32B, which followed the official recommended settings: a temperature of 1.0, a top-$p$ value of 0.7, and a maximum generation length of 20,480 tokens. For AIME2024 and AIME2025, we report accuracy by averaging over 16 sampled generations per question, while for MATH500 and GPQA, accuracy is computed using a single sampled generation per question.

% \section{SFT vs RL}

\section{Additional Experiments and Analysis}
\subsection{Matched SFT-vs.-GRPO Comparison}
\label{appendix:sft_vs_grpo}

To complement the observational comparison in Section~\ref{sec:observation_study}, we conduct a matched comparison using the same Qwen2.5-7B-Instruct backbone and the same multilingual MATH500 evaluation protocol. We compare the initial model, an SFT model trained on 1K English MATH examples, and a GRPO model trained on the same 1K English MATH examples.

Table~\ref{tab:appendix_sft_vs_grpo} shows that GRPO achieves the strongest multilingual performance under this controlled setting. Compared with SFT, GRPO improves average accuracy from 50.80 to 54.24, non-English average accuracy from 48.20 to 51.74, and low-resource average accuracy from 22.13 to 29.53. In contrast, SFT improves English accuracy over the base model but does not consistently improve multilingual performance, and it increases the average off-target rate.

This comparison supports the observational trend in Figure~\ref{fig:open_source_model_MTI} under a matched backbone and evaluation setting. However, it is limited to one backbone, one training set, and one benchmark, and should therefore not be interpreted as a universal claim that GRPO or RL-style post-training always outperforms SFT.

\begin{table}[t]
\centering
\small
\setlength{\tabcolsep}{4.5pt}
\begin{tabular}{lrrrrr}
\toprule
\textbf{Method} & \textbf{En} & \textbf{Avg} & \textbf{Non-En} &
\textbf{Low-res.} & \textbf{Off-tgt.} \\
\midrule
Base & 74.80 & 50.91 & 48.52 & 27.40 & 0.18 \\
SFT  & 76.80 & 50.80 & 48.20 & 22.13 & 4.40 \\
GRPO & 79.20 & 54.24 & 51.74 & 29.53 & 0.49 \\
\bottomrule
\end{tabular}
\caption{\textbf{Matched SFT-vs.-GRPO comparison on multilingual MATH500.}
All values are percentages. All models use Qwen2.5-7B-Instruct as the backbone. SFT and GRPO are trained on the same 1K English MATH examples and evaluated under the same multilingual MATH500 protocol. Low-res. denotes average accuracy over \emph{bn}, \emph{sw}, and \emph{te}; Off-tgt. denotes the off-target language rate.}
\label{tab:appendix_sft_vs_grpo}
\end{table}

\subsection{Absolute-Transfer Metrics under Parallel Training}
\label{appendix:absolute_transfer_metrics}

Since MTI measures relative transfer efficiency rather than absolute multilingual reasoning ability, we additionally report several absolute-transfer metrics for the Qwen2.5-7B-Instruct parallel-training experiments. Specifically, we compute non-English average accuracy, which excludes English from the language average; the English--non-English gap, which measures how far non-English performance lags behind English; and low-resource average accuracy over \emph{bn}, \emph{sw}, and \emph{te}. MTI is retained only as an auxiliary diagnostic of relative transfer efficiency.

\begin{table}[t]
\centering
\small
\setlength{\tabcolsep}{3.5pt}
\begin{tabular}{lrrrrr}
\toprule
\textbf{Setting} & \textbf{En} & \textbf{Non-En} & \textbf{Gap} &
\textbf{Low-res.} & \textbf{MTI} \\
\midrule
English-only & 79.20 & 51.74 & 27.46 & 29.53 & 1.163 \\
+1 parallel  & 78.40 & 55.82 & 22.58 & 36.27 & 2.496 \\
+2 parallels & 79.00 & 56.32 & 22.68 & 34.67 & 2.650 \\
+3 parallels & 77.80 & 56.44 & 21.36 & 34.73 & 3.002 \\
+4 parallels & 77.20 & 56.76 & 20.44 & 37.13 & 3.282 \\
+5 parallels & 77.40 & 57.18 & 20.22 & 39.33 & 3.475 \\
+6 parallels & 76.40 & 57.72 & 18.68 & 40.53 & 3.534 \\
+7 parallels & 76.60 & 57.81 & 18.79 & 40.47 & 3.631 \\
\bottomrule
\end{tabular}
\caption{\textbf{Absolute-transfer metrics for Qwen2.5-7B-Instruct under parallel training.}
All accuracy and gap values are percentages. Non-En denotes average accuracy over non-English languages; Gap denotes the En--Non-En accuracy gap; Low-res. denotes average accuracy over \emph{bn}, \emph{sw}, and \emph{te}. MTI is reported as an auxiliary relative transfer-efficiency metric.}
\label{tab:absolute_transfer_metrics}
\end{table}

Table~\ref{tab:absolute_transfer_metrics} shows that the benefit of parallel multilingual supervision is visible not only in MTI, but also in absolute multilingual performance. Moving from English-only training to one added parallel language improves non-English average accuracy from 51.74 to 55.82, reduces the English--non-English gap from 27.46 to 22.58, and improves low-resource average accuracy from 29.53 to 36.27. Adding more parallel languages further improves non-English average accuracy to 57.81 and reduces the English--non-English gap to 18.79 with seven added languages.

These results support the interpretation that MTI should be read together with absolute performance metrics. A high MTI indicates strong relative transfer efficiency, but it does not by itself imply high absolute multilingual reasoning accuracy. We therefore use MTI as a complementary diagnostic rather than as the sole measure of cross-lingual reasoning transfer.

\subsection{Functional-Form Comparison and Marginal-Gain Analysis}
\label{appendix:scaling_fit_analysis}

In Section~\ref{sec:parallel_scaling_law}, we use a power-law-like curve to summarize the diminishing-return trend in the multilingual regime. This curve is intended only as a descriptive fit, not as a theoretical scaling law. To examine whether this descriptive choice is reasonable, we compare four candidate functional forms on the multilingual regime, where the number of training languages is $X=2,\ldots,8$:
\begin{align}
\text{linear:}\quad & f(X)=aX+b, \\
\text{logarithmic:}\quad & f(X)=a\log X+b, \\
\text{power-law:}\quad & f(X)=aX^b, \\
\text{saturation:}\quad & f(X)=c-ae^{-bX}.
\end{align}
We evaluate the fits using $R^2$, leave-one-out cross-validation mean squared error (LOOCV MSE), Akaike information criterion (AIC), and Bayesian information criterion (BIC). LOOCV MSE measures out-of-sample predictive error, while AIC and BIC compare fit quality with penalties for model complexity.

\begin{table}[t]
\centering
\small
\setlength{\tabcolsep}{8pt}
\begin{tabular}{lrrrr}
\toprule
\textbf{Fit} & $\mathbf{R^2}$ & \textbf{MSE} & \textbf{AIC} & \textbf{BIC} \\
\midrule
\multicolumn{5}{l}{\textbf{Qwen MTI}} \\
Linear     & 0.951 & 0.0178 & -29.523 & -29.631 \\
Log        & 0.974 & 0.0121 & -33.950 & -34.058 \\
Power-law  & 0.977 & 0.0082 & -34.653 & -34.762 \\
Saturation & 0.982 & 0.0894 & -34.368 & -34.530 \\
\midrule
\multicolumn{5}{l}{\textbf{Llama MTI}} \\
Linear     & 0.973 & 0.0015 & -46.175 & -46.284 \\
Log        & 0.981 & 0.0012 & -48.701 & -48.809 \\
Power-law  & 0.984 & 0.0010 & -49.706 & -49.814 \\
Saturation & 0.000 & 0.0388 & -18.912 & -19.075 \\
\bottomrule
\end{tabular}
\caption{\textbf{Curve-fitting comparison for MTI in the multilingual regime.}
Fits are evaluated on configurations with $X=2,\ldots,8$. MSE denotes the leave-one-out cross-validation (LOOCV) MSE.}
\label{tab:appendix_curve_fitting_comparison}
\end{table}

As shown in Table~\ref{tab:appendix_curve_fitting_comparison}, the power-law fit achieves the best LOOCV MSE, AIC, and BIC for both Qwen and Llama MTI, with competitive $R^2$. We therefore use the power-law-like curve as a compact descriptive summary of the multilingual regime, rather than as evidence for a universal scaling law.

We also examine the role of the monolingual point $X=1$ under an all-point fit. As an influence diagnostic, we fit the power-law curve on all observed points $X=1,\ldots,8$ and compute Cook's distance for each point.

\begin{table}[t]
\centering
\small
\setlength{\tabcolsep}{3.8pt}
\begin{tabular}{crrrr}
\toprule
$\mathbf{X}$ &
\textbf{Qwen MTI} &
\textbf{Qwen Acc.} &
\textbf{Llama MTI} &
\textbf{Llama Acc.} \\
\midrule
1 & 2.005 & 4.503 & 3.684 & 3.341 \\
2 & 0.414 & 0.512 & 0.495 & 0.862 \\
3 & 0.015 & 0.099 & 0.067 & 0.004 \\
4 & 0.019 & 0.001 & 0.005 & 0.015 \\
5 & 0.016 & 0.007 & 0.009 & 0.000 \\
6 & 0.009 & 0.009 & 0.008 & 0.000 \\
7 & 0.029 & 0.008 & 0.064 & 0.005 \\
8 & 0.284 & 0.072 & 0.068 & 0.029 \\
\bottomrule
\end{tabular}
\caption{\textbf{Cook's distance under all-point power-law fits.}
The power-law curve is fit to all observed points, $X=1,\ldots,8$.}
\label{tab:appendix_cooks_distance}
\end{table}

Table~\ref{tab:appendix_cooks_distance} shows that $X=1$ has the largest Cook's distance in all four settings. It also exceeds the common $4/n$ rule-of-thumb threshold, which is $0.5$ for $n=8$. We treat this result as an influence diagnostic: it suggests that the monolingual point is influential under an all-point power-law fit, but it is not a formal statistical test of a separate regime.

Finally, we complement the curve-fitting analysis with a direct marginal-gain analysis. Table~\ref{tab:appendix_marginal_gain_decomposition} reports the gain from each consecutive increase in the number of training languages.

\begin{table}[t]
\centering
\small
\begin{tabular}{lrrrr}
\toprule
\textbf{Transition} & \textbf{Avg Acc} & \textbf{Non-En} & \textbf{Low-res.} & \textbf{MTI} \\
\midrule
$1\rightarrow2$ & +3.70 & +4.08 & +6.73 & +1.333 \\
$2\rightarrow3$ & +0.50 & +0.50 & -1.60 & +0.154 \\
$3\rightarrow4$ & +0.00 & +0.12 & +0.07 & +0.352 \\
$4\rightarrow5$ & +0.20 & +0.32 & +2.40 & +0.280 \\
$5\rightarrow6$ & +0.40 & +0.42 & +2.20 & +0.193 \\
$6\rightarrow7$ & +0.40 & +0.54 & +1.20 & +0.059 \\
$7\rightarrow8$ & +0.10 & +0.09 & -0.07 & +0.097 \\
\bottomrule
\end{tabular}
\caption{\textbf{Marginal-gain decomposition over consecutive training-language counts.}
Each row reports the gain from increasing the number of training languages from $X$ to $X+1$ in the Qwen2.5-7B-Instruct parallel-training experiments.}
\label{tab:appendix_marginal_gain_decomposition}
\end{table}

Table~\ref{tab:appendix_marginal_gain_decomposition} shows that the $1\rightarrow2$ transition yields the largest consecutive gain across average accuracy, non-English average accuracy, low-resource average accuracy, and MTI. Additional languages continue to provide gains on most metrics, but the marginal improvements are smaller.

We further summarize how much of the total $X=1\rightarrow8$ gain is explained by the first added parallel language in Table~\ref{tab:appendix_first_parallel_gain}.

\begin{table}[t]
\centering
\small
\setlength{\tabcolsep}{10pt}
\begin{tabular}{lrrr}
\toprule
\textbf{Metric} & $\mathbf{1\rightarrow2}$ & $\mathbf{1\rightarrow8}$ & \textbf{Fraction} \\
\midrule
Avg Acc.      & +3.70  & +5.30  & 69.8\% \\
Non-En Acc.   & +4.08  & +6.07  & 67.2\% \\
Low-res. Acc. & +6.73  & +10.93 & 61.6\% \\
MTI           & +1.333 & +2.468 & 54.0\% \\
\bottomrule
\end{tabular}
\caption{\textbf{Fraction of total gain explained by the first added parallel language.}
The table compares the gain from English-only to bilingual training
($1\rightarrow2$) with the total gain from English-only to eight-language
training ($1\rightarrow8$). Avg Acc. denotes average accuracy over all
languages; Non-En Acc. denotes average accuracy over non-English languages;
Low-res. Acc. denotes average accuracy over \emph{bn}, \emph{sw}, and \emph{te}.}
\label{tab:appendix_first_parallel_gain}
\end{table}

Together, Tables~\ref{tab:appendix_marginal_gain_decomposition} and~\ref{tab:appendix_first_parallel_gain} show that the English-only to bilingual transition is the largest observed consecutive improvement and explains a large fraction of the total improvement from $X=1$ to $X=8$: 69.8\% for average accuracy, 67.2\% for non-English average accuracy, 61.6\% for low-resource average accuracy, and 54.0\% for MTI. Additional languages continue to provide smaller but generally positive gains.

\subsection{Parallel vs.\ Non-Parallel Multilingual Supervision}
\label{appendix:parallel_vs_nonparallel}

To further examine whether the bilingual gain comes from same-problem cross-lingual alignment rather than multilingual exposure alone, we compare parallel bilingual supervision with non-parallel multilingual supervision. We compute the advantage of parallel training as
\begin{equation}
\text{Advantage} = Acc_{\text{parallel}} - Acc_{\text{non-parallel}}
\end{equation}
for each benchmark-language pair.

\begin{table}[t]
\centering
\small
\renewcommand\arraystretch{1.0}
\setlength{\tabcolsep}{8pt}
\begin{tabular}{lrrr}
\toprule
\textbf{Group} & $\mathbf{n}$ & \textbf{Mean Adv.} & \textbf{W/T/L} \\
\midrule
Overall          & 44 & +0.995 & 28 / 8 / 8 \\
Non-English      & 40 & +1.101 & 28 / 5 / 7 \\
Unseen Non-En    & 36 & +1.213 & 27 / 5 / 4 \\
Unseen low-res.  & 12 & +1.632 & 9 / 3 / 0 \\
Added lang. (ru) & 4  & +0.093 & 1 / 0 / 3 \\
English          & 4  & -0.063 & 0 / 3 / 1 \\
\bottomrule
\end{tabular}
\caption{\textbf{Parallel vs.\ non-parallel multilingual supervision.}
The advantage is computed as
$Acc_{\text{parallel}} - Acc_{\text{non-parallel}}$
over benchmark-language pairs. W/T/L denotes wins, ties, and losses.}
\label{tab:appendix_parallel_vs_nonparallel}
\end{table}

Table~\ref{tab:appendix_parallel_vs_nonparallel} shows that parallel supervision improves accuracy by 0.995 points on average across all benchmark-language pairs and wins on 28 out of 44 pairs. The advantage is larger for unseen non-English languages and unseen low-resource languages, while the added language itself shows only a small mean advantage. This suggests that the gain is not merely localized to the added parallel language, but transfers to other unseen languages as well.

\subsection{Additional Analysis of Open-Source Model Transfer}
\label{appendix:open_source_model_analysis}

We provide additional examples to clarify how Figure~\ref{fig:open_source_model_MTI} supports the observation that cross-lingual transfer is highly variable across models, training paradigms, and target languages.

First, strong English performance does not necessarily imply strong multilingual performance. For example, on MATH500, DeepSeek-R1-Distill-Qwen-7B achieves 86.2\% accuracy on English, but only 18.4\% on Swahili and 17.6\% on Telugu. Similarly, OpenThinker2-7B reaches 86.0\% on English, but only 12.6\% on Swahili and 8.4\% on Telugu.

Second, transfer can differ substantially even within the same model family. For example, Qwen-2.5-7B-SimpleRL-Zoo and Qwen-2.5-14B-SimpleRL-Zoo are from the same model family, but their MATH500 average accuracies differ substantially, 56.47\% versus 63.93\%. Their low-resource performance also changes noticeably; for example, Telugu accuracy increases from 25.8\% to 41.0\%.

Third, Figure~\ref{fig:open_source_model_MTI} suggests an observational training-paradigm trend: RL-style models have higher average MTI than SFT-style models in our open-source model pool. However, since these models differ in backbone, data, and training recipe, this pattern should be interpreted as observational rather than causal. We therefore use this analysis as motivation for the controlled comparisons in Section~\ref{sec:interventional_study}.

\subsection{Language-Level Accuracy and Relative Variance}
\label{appendix:language_variance_analysis}

We further analyze whether languages with lower average accuracy also exhibit larger relative variation across models. For each benchmark and language, we compute the mean accuracy and coefficient of variation (CV) across open-source models:
\begin{equation}
\mathrm{CV}=\frac{\mathrm{std}}{\mathrm{mean}}.
\end{equation}
We use CV instead of raw standard deviation to account for differences in average accuracy across languages.

\begin{table}[t]
\centering
\small
\setlength{\tabcolsep}{10pt}
\begin{tabular}{lrr}
\toprule
\textbf{Scope} & \textbf{Pearson $r$} & \textbf{Spearman $\rho$} \\
\midrule
Overall      & -0.956 & -0.882 \\
MATH500      & -0.971 & -0.946 \\
AIME24       & -0.901 & -0.946 \\
AIME25       & -0.948 & -0.936 \\
GPQA-Diamond & -0.777 & -0.727 \\
\bottomrule
\end{tabular}

\caption{\textbf{Correlation between mean accuracy and relative variance across languages.}
For each benchmark and language, mean accuracy and CV are computed across open-source models. Negative correlations indicate that lower-accuracy languages tend to have higher relative variance.}
\label{tab:appendix_language_variance}
\end{table}

Table~\ref{tab:appendix_language_variance} shows a strong negative correlation between language-level mean accuracy and CV across benchmarks. This indicates that lower-accuracy languages tend to exhibit higher relative variation across models. At the group level, low-resource languages (\emph{bn}, \emph{sw}, and \emph{te}) also have a higher mean CV than the other languages (0.956 vs.\ 0.706). These results suggest that the observed target-language variation reflects not only language identity or resource level, but also higher relative instability in low-accuracy regimes.

\section{Detailed Results and Analysis}
\subsection{Robustness of Transfer Metrics}
\label{appendix:mti_robustness}

To examine whether our conclusions depend on the specific choice of MTI, we repeated the main analyses using two alternative transfer metrics.

First, we consider a normalized-gain variant:
\begin{equation}
\Delta N_{b,l} = \frac{Acc^{\text{trained}}_{b,l} - Acc^{\text{base}}_{b,l}}{1 - Acc^{\text{base}}_{b,l}},
\end{equation}
and define the corresponding transfer ratio by replacing $\Delta R$ with $\Delta N$ in the MTI computation.

Second, we consider a raw-gain ratio:
\begin{equation}
\Delta G_{b,l} = Acc^{\text{trained}}_{b,l} - Acc^{\text{base}}_{b,l},
\end{equation}
and compare the gain on an unseen language with the average gain on the training language(s):
\begin{equation}
\text{RG}_{b,l_{\text{unseen}}} = \frac{\Delta G_{b,l_{\text{unseen}}}}{\Delta G_{b,\mathcal{L}_{\text{train}}}}.
\end{equation}

Table~\ref{tab:metric_robustness_summary} summarizes the corresponding results on the main parallel-training comparison. Although the absolute values differ across metrics, the main qualitative pattern remains unchanged: moving from English-only to bilingual training yields the largest gain, and adding another parallel language provides a smaller additional improvement. This supports the main-text conclusion that the First-Parallel Leap is not specific to the original MTI definition.

\input{tabs/appendix_12_robustness_of_MTI}

\paragraph{Sensitivity to base accuracy.}
Because MTI is based on relative gain, it can become sensitive when base accuracy is low. To assess whether the main result is driven by such cases, we recompute the transfer scores on multilingual MATH500 after excluding unseen-language cases whose base accuracy falls below different thresholds. Table~\ref{tab:mti_base_sensitivity} shows that the absolute values vary across metrics and filters, especially for the original relative-gain MTI. However, the main qualitative result remains unchanged: the transition from monolingual to bilingual training yields a clear improvement under all transfer metrics and filtering thresholds. This suggests that the bilingual gain is not an artifact of the lowest-base-accuracy cases alone.

\input{tabs/appendix_11_sensitivity_to_base_acc}

\subsection{Observational Study}
\label{appendix:observation_study}
\subsubsection{Template Choice for Base Models}

To evaluate base models consistently, we compare three template settings for \textit{Qwen2.5-7B-Base} and \textit{Qwen2.5-Math-7B-Base}: \texttt{Qwen-Math Template}, \texttt{Qwen-Instruct Template}, and \texttt{No Template}. As shown in Table~\ref{tab:appendix4_base_model_template}, the \texttt{Qwen-Instruct Template} yields the best overall reasoning accuracy and target-language consistency on multilingual MATH500. We therefore use it as the default template for evaluating all general-base and math-base models.

\input{tabs/appendix_4_base_model_template}

\subsubsection{Detailed Results for Initial Model Types}

Table~\ref{tab:appendix5_initial_model_acc_and_off-tag} reports the full multilingual accuracy and off-target results for the initial models used in our analysis. Two patterns are worth noting. First, within the Qwen2.5-7B series, the instruction-tuned model has the lowest off-target rate and the highest final multilingual accuracy. Second, within the Qwen2.5-Base series, larger models generally achieve higher multilingual accuracy and lower off-target rates. These results support the main-text observation that stronger final multilingual performance and stronger relative transfer efficiency are not always aligned.

\input{tabs/appendix_5_observational_study_initial_model_acc_and_off-tag}

\subsubsection{Detailed Results for Open-source Models}

Table~\ref{tab:appendix7_open_models_acc_and_off-tag} provides the full transfer, accuracy, and off-target results for the open-source models in the observational study. These tables complement Figure~\ref{fig:open_source_model_MTI} in the main text by showing the benchmark-level and language-level variation behind the averaged results.

% \input{tabs/appendix_6_observational_study_open_source_models_TI}
% \input{tabs/appendix_7_observational_study_open_models_acc_and_off-tag}

% \begin{figure*}[!h]
% \centering
% \begin{subfigure}{0.495\linewidth}
%     \centering
%     \includegraphics[width=\linewidth]{figs/sft_tuned_models_across_languages.pdf}
%     \caption{SFT-tuned Models}
%     \label{fig:sft_tuned_models_across_languages}
% \end{subfigure}
% \begin{subfigure}{0.495\linewidth}
%     \centering
%     \includegraphics[width=\linewidth]{figs/rl_tuned_models_across_languages.pdf}
%     \caption{RL-tuned Models}
%     \label{fig:rl_tuned_models_across_languages}
% \end{subfigure}

% \caption{\textbf{The Performance of Various Open-source Models. Part 2:} The transferability difference between SFT-tuned and RL-tuned models across languages. Note that ``\textit{None}'' values indicate that \textit{Qwen-2.5-1.5B} achieved zero accuracy in most languages on AIME24 and AIME25, making relative gain undefined.}
% \label{fig:open_models_between_sft_and_rl_tuned_models_across_languages}
% \end{figure*}

\input{tabs/appendix_10_the_impact_of_hyperparameter_lambda}

\subsection{Controlled Comparative Study}

\subsubsection{Reward Hyperparameter Sensitivity}
\label{sec:impact_of_reward_hyperpameters}

To examine the role of different reward components, we vary the weights on the accuracy, format, and language-consistency rewards. Table~\ref{tab:appendix_10_the_impact_of_hyperparameter_lambda} reports the corresponding results. Performance is most sensitive to the accuracy reward and the format reward. Removing the language-consistency reward substantially increases the off-target rate, while removing the format reward leads to the largest overall degradation. We use this analysis as a sensitivity check on the reward design; it does not fully disentangle reasoning improvement from formatting or language compliance.

\subsubsection{Detailed Results for Initial Model Types}
\label{appendix:the_impact_of_initial_model_types}

Table~\ref{tab:experiment_2_control_study_initial_model} reports the full results for the initial model type comparison in the controlled comparative study.

\input{tabs/experiment_2_control_study_initial_model}

\subsubsection{Detailed Results for Model Family}
\label{appendix:the_impact_of_different_model_families}

\begin{figure*}[!h]
\centering
\begin{subfigure}{\linewidth}
    \centering
    \includegraphics[width=0.95\linewidth]{figs/impact_of_different_initial_model_qwen_vs_llama_on_math500_v2.pdf}
    \caption{MATH500}
\end{subfigure}
\begin{subfigure}{\linewidth}
    \centering
    \includegraphics[width=0.95\linewidth]{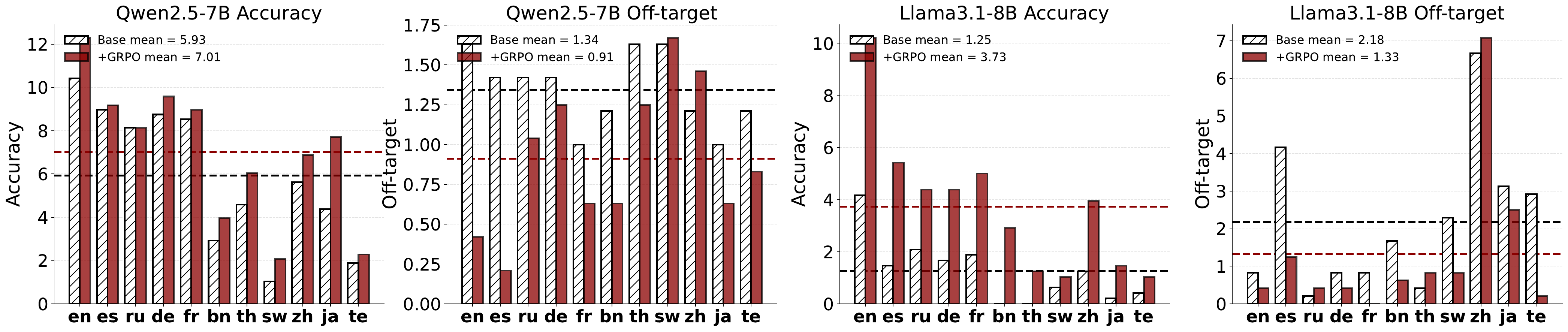}
    \caption{AIME24}
\end{subfigure}

\begin{subfigure}{\linewidth}
    \centering
    \includegraphics[width=0.95\linewidth]{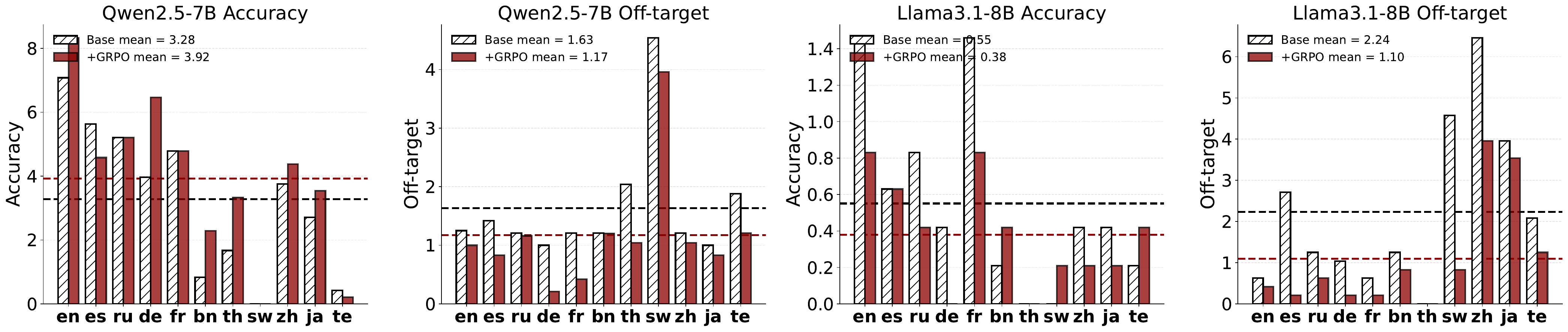}
    \caption{AIME25}
\end{subfigure}

\begin{subfigure}{\linewidth}
    \centering
    \includegraphics[width=0.95\linewidth]{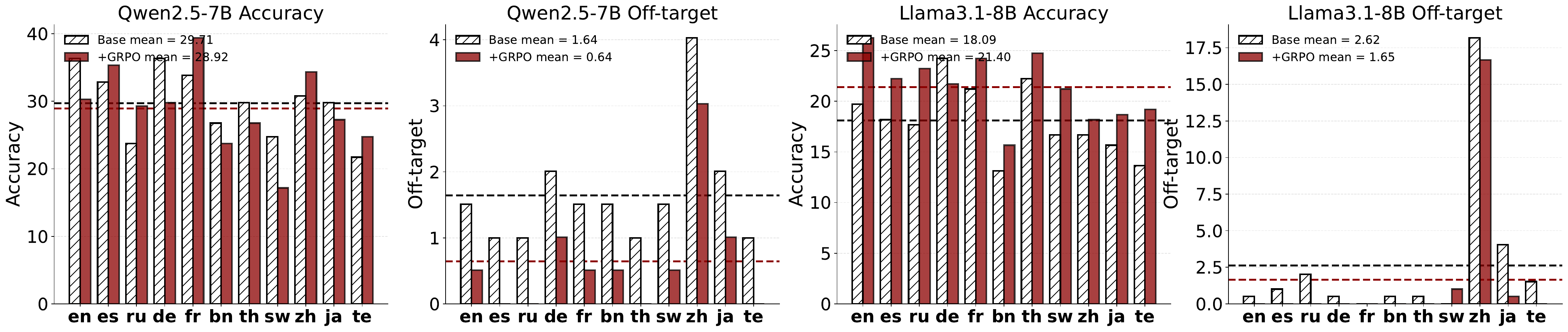}
    \caption{GPQA-Diamond}
\end{subfigure}

\caption{\textbf{The Impact of Different Model Families in Controlled Comparative Study.} Multilingual reasoning performance across languages, comparing the influence of model family using Qwen2.5-7B-Instruct and Llama3.1-8B-Instruct as initial models. 
 }
\label{fig:controlled_exp_different_initial_model_qwen_vs_llama_all_benchmark}
\end{figure*}

Figure~\ref{fig:controlled_exp_different_initial_model_qwen_vs_llama_all_benchmark} provides the full benchmark-level results for the model family comparison between Qwen2.5-7B-Instruct and Llama3.1-8B-Instruct. In both families, English-only GRPO improves performance across languages, but the relative gains differ substantially. As discussed in the main text, this comparison should be interpreted as a controlled comparison under matched post-training settings rather than as a pure estimate of family effects, since tokenization, pretraining data, and architecture remain entangled.

%%%%%
\paragraph{Cross-lingual Transfer within the Sino-Tibetan Family.}

To complement the model family comparison, we also evaluate Qwen2.5-7B-Instruct and Meta-Llama-3.1-8B-Instruct on several Sino-Tibetan languages after English-only GRPO. Table~\ref{tab:sino_tibetan} shows that Qwen retains higher absolute accuracy in Chinese, while Llama exhibits larger relative gains on lower-performing languages such as Tibetan and Myanmar. We report this result as an additional case study consistent with the main-text observation that stronger starting performance and larger relative gains do not always coincide.

\input{tabs/appendix_13_sino_tibetan_family}

\subsubsection{Detailed Results for Model Size}
\label{appendix:the_impact_of_model_size}

Table~\ref{tab:appendix8_control_strudy_scaling_up} reports the full results for the model size comparison between \textit{Qwen2.5-1.5B-Instruct} and \textit{Qwen2.5-7B-Instruct}. The smaller model shows larger gains on multilingual MATH500 and GPQA-Diamond, whereas the larger model attains stronger final performance on the harder AIME benchmarks. These results support the main-text observation that the effect of model scale depends on the interaction between initial model strength and benchmark difficulty.

\input{tabs/appendix_8_control_study_scaling_up}

\subsection{Parallel Training Results}
\label{appendix:parallel_scaling_law}

\subsubsection{Detailed Results of Parallel Training}

Table~\ref{tab:appendix2_scaling_parallel_law_acc} reports the full accuracy results across languages for different numbers of training languages. These tables complement Figure~\ref{fig:parallel_exp_scaling_law} in the main text by providing the full language-level breakdown behind the averaged trends.

\input{tabs/appendix_2_scaling_parallel_law_acc}

\subsubsection{Parallel Training Language Settings}

Table~\ref{tab:appendix9_scaling_parallel_law_languages} lists the language configurations used in the parallel training study, where the number of training languages increases from 1 to 7.

\input{tabs/appendix_9_scaling_parallel_law_languages}

\subsubsection{Effect of the Selected Parallel Language}
\label{appendix:the_impact_of_selected_languages}

\begin{figure*}[!h]
\centering
\begin{subfigure}{\linewidth}
    \centering
    \includegraphics[width=0.95\linewidth]{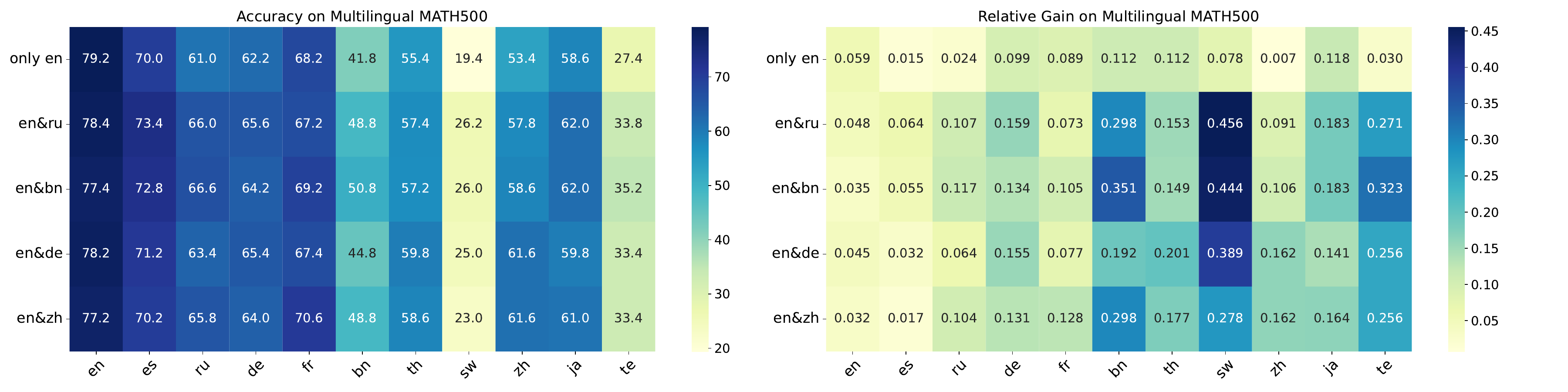}
    \caption{MATH500}
    \label{fig:parallel_with_different_langs_math500}
\end{subfigure}
\begin{subfigure}{\linewidth}
    \centering
    \includegraphics[width=0.95\linewidth]{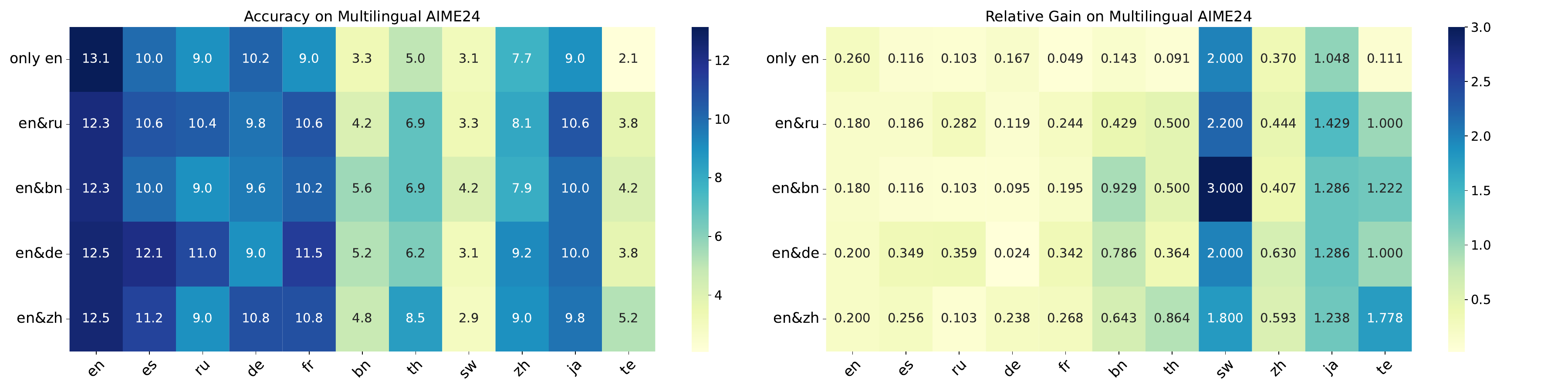}
    \caption{AIME24}
    \label{fig:parallel_with_different_langs_aime24}
\end{subfigure}

\begin{subfigure}{\linewidth}
    \centering
    \includegraphics[width=0.95\linewidth]{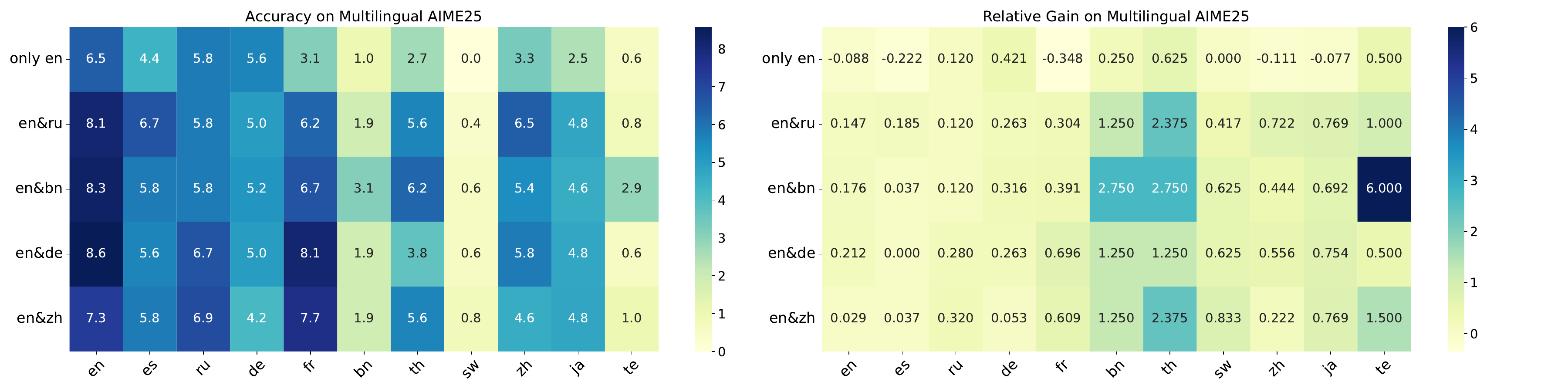}
    \caption{AIME25}
    \label{fig:parallel_with_different_langs_aime25}
\end{subfigure}

\begin{subfigure}{\linewidth}
    \centering
    \includegraphics[width=0.95\linewidth]{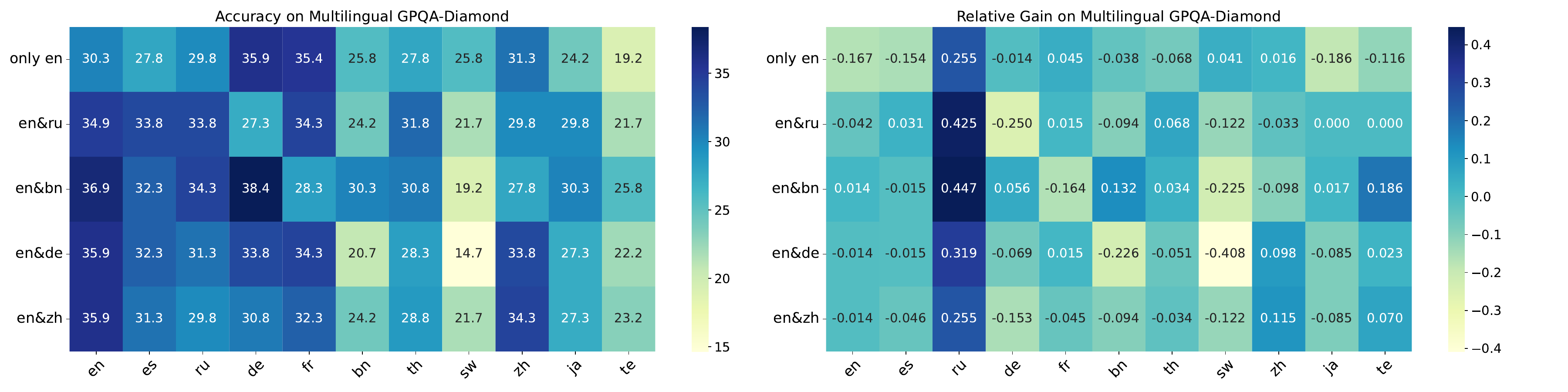}
    \caption{GPQA-Diamond}
    \label{fig:parallel_with_different_langs_gpqa}
\end{subfigure}

\caption{\textbf{Analysis across Selected Parallel Languages.} The accuracy and relative gain across various benchmarks with different parallel languages. ``Only en'' denotes only fine-tuned on English data. ``en\&\texttt{LANGUAGE}'' indicates the model was fine-tuned on English and a parallel language, with \texttt{LANGUAGE} representing \emph{ru, bn, de, zh}, respectively.}
\label{fig:parallel_with_different_langs}
\end{figure*}

Figure~\ref{fig:parallel_with_different_langs} reports benchmark-level results for different choices of the added parallel language. Across settings, the same qualitative pattern remains: adding one parallel language consistently improves transfer, and the differences across language choices are modest relative to the monolingual-to-bilingual jump. Some language-specific effects remain, especially in lower-resource languages, but the main bilingual gain is robust to the choice of the added language.

\subsection{Parallel Training from a Chinese-Centric Perspective}
\label{subsec:chinese_centric}

Table~\ref{tab:chinese_parallel_scaling} reports the full results for three settings: monolingual Zh training, bilingual Zh+Ru training, and trilingual Zh+Ru+Fr training. A similar qualitative pattern appears: adding one parallel language yields the largest improvement in average accuracy and MTI, while adding a second language provides a smaller additional gain. We report this result as supporting evidence that the bilingual gain is not unique to English-only post-training.

\input{tabs/appendix_12_chinese-centric}

\section{Dataset License}
We use publicly available models and benchmarks, including the multilingual versions of MATH500~\citep{hendrycks2021measuring}, AIME2024, AIME2025 from XReasoning~\citep{qi2025models}, and GPQA-Diamond~\citep{rein2023gpqa} from BenchMAX~\citep{huang2025benchmax}, and follow their corresponding usage guidelines.

\input{tabs/appendix_7_observational_study_open_models_acc_and_off-tag}

\clearpage
\onecolumn
\section{Prompts Template}
\subsection{Templates for Base Model}
\label{sec:template_for_base_models}
\input{prompt/template_for_base_models}

\subsection{Multilingual Reasoning Instruction}
\input{prompt/instruction_used_for_open-source_model}

\subsection{Prompt hacking to force response language}
\label{appendix:prompt_hacking}
\input{prompt/prompt_hack}

\subsection{Template for R1-like Reasoning}
\input{prompt/template_for_r1_like_reasoning}

% \section{Takeaway: Synthesis of Findings}
% \input{takeaway/overview_of_takeaway}

\end{document}

%% file: tabs/experiment_2_control_study_initial_model_simplify.tex
\begin{table}[!h]
\footnotesize
\renewcommand\arraystretch{1.2}
\setlength{\tabcolsep}{2.2mm}

\begin{center}
\begin{tabular}{lccc}
\toprule[1.2pt] 
\textbf{Model} & \textbf{Acc} & \textbf{Off-target} & \textbf{MTI}  \\ 
\midrule[0.8pt] 
\rowcolor{lightgray} 
Qwen2.5-7B-Base  & 12.00 & 11.41 & - \\
\hspace{8pt} $\looparrowright$ \textit{GRPO on En Data} &22.45	&3.12	&1.95 \\
\rowcolor{lightgray} 
Qwen2.5-7B-Instruct & 22.45 & 1.43 & - \\
\hspace{8pt} $\looparrowright$ \textit{GRPO on En Data} & 23.51 & 0.94 & 1.23 \\
\rowcolor{lightgray} 
Qwen2.5-Math-7B & 12.25 & 22.59 & - \\
\hspace{8pt} $\looparrowright$ \textit{GRPO on En Data}	&19.37	&9.50	&2.12 \\
% \rowcolor{lightgray} 
% Meta-Llama-3.1-8B-Instruct & 27.07 & 1.25 & 0.55 & 18.09 & 11.74 & 2.33 & - \\
% \hspace{8pt} $\looparrowright$ \textit{GRPO on En Data} &  36.11 & 3.73 & 0.38 & 21.40 & 15.40 & 1.34 & 1.72 \\

\bottomrule[1.2pt]
\end{tabular}
\caption{\textbf{The Impact of Initial Model Type.} Average Accuracy (\%), Off-target rate (\%), and MTI across eleven languages on four benchmarks.}

\label{tab:experiment_2_control_study_initial_model_simplify}
\end{center}
\vspace{-6mm}
\end{table}

%% file: tabs/experiment_4_fixed_budget.tex
\begin{table}[!h]
\small
\renewcommand\arraystretch{1.2}
\setlength{\tabcolsep}{0.7mm}

\begin{center}
\begin{tabular}{lcccc}
\toprule[1.2pt] 
\textbf{Budget} & \textbf{Type} & \textbf{Acc} & \textbf{Off-target} & \textbf{MTI}  \\ 
\midrule[0.8pt] 
\rowcolor{lightgray} 
$1\times$& $1\times$En        & 54.24 & 0.49 & 1.16 \\
\midrule[0.8pt] 
$2\times$& $2\times$En        & 55.82 & 0.58 & 1.92 \\
$2\times$& $1\times$En + $1\times$ Ru       & 57.87 & 0.20 & 2.50 \\
\midrule[0.8pt] 
$3\times$& $3\times$En        & 57.13 & 0.42 & 2.38 \\
$3\times$& $1\times$En + $1\times$ Ru + $1\times$ Fr     & 58.38 & 0.24 & 2.65 \\

\bottomrule[1.2pt]
\end{tabular}
\caption{The Fixed-Budget ablation experiment based on Qwen2.5-7B-Instruct.}
\label{tab:experiment_4_fixed_budget}
\end{center}
\vspace{-6mm}
\end{table}

%% file: tabs/experiment_7_prompt_ablation.tex
\begin{table}[!h]
\centering
\small
\renewcommand\arraystretch{1.2}
\setlength{\tabcolsep}{1.5mm}
\begin{tabular}{llcc}
\toprule[1.2pt]
\textbf{Evaluation} & \textbf{Training} & \textbf{Acc} & \textbf{Off-target} \\
\midrule[0.8pt]
No language control & English-only & 57.82 & 75.76 \\
No language control & English+Ru & 58.43 & 0.36 \\
Language-controlled & English-only & 54.20 & 0.50 \\
Language-controlled & English+Ru & 57.90 & 0.20 \\
\bottomrule[1.2pt]
\end{tabular}
\caption{\textbf{Prompt ablation.} Average accuracy and off-target rate on multilingual MATH500 under two evaluation protocols.}
\vspace{-4mm}
\label{tab:measurement_disentanglement_prompt}
\end{table}

%% file: tabs/experiment_8_chinese_centric.tex
\begin{table}[!h]
\centering
\small
\renewcommand\arraystretch{1.2}
\setlength{\tabcolsep}{1.5mm}
\begin{tabular}{llll}
\toprule[1.2pt]
\textbf{Training Data} & \textbf{Acc} & \textbf{Off-target} & \textbf{MTI} \\
\midrule[0.8pt]
Zh & 55.00 & 0.38 & 0.86 \\
Zh + Ru & 58.13$_{(+3.13)}$ & 0.26$_{(-0.12)}$ & 1.69$_{(+0.83)}$ \\
Zh + Ru + Fr & 58.76$_{(+3.76)}$ & 0.23$_{(-0.15)}$ & 1.87$_{(+1.01)}$ \\
\bottomrule[1.2pt]
\end{tabular}
\caption{\textbf{Chinese-centric parallel training results} on multilingual MATH500.}
\label{tab:chinese_parallel_scaling_main}
\vspace{-4mm}
\end{table}

%% file: tabs/appendix_1_open_source_model.tex
\begin{table*}[h]
    \centering
    \small
    \renewcommand\arraystretch{1.2}
    \setlength{\tabcolsep}{2.5mm}
    \begin{tabular}{llcc}
        \toprule[1.2pt]
        \textbf{Model} & \textbf{Initial Model} & \textbf{Size} & \textbf{Training Paradigm} \\
        \midrule[0.8pt] 
        \midrule[0.8pt]
        \rowcolor{lightblue} 
        DeepSeek-R1-Distill-Qwen-7B & Qwen2.5-Math-7B-Base & 7B & SFT \\
        \rowcolor{lightpink} 
        Open-Reasoner-Zero-7B & Qwen2.5-7B-Base & 7B & RL \\
        \rowcolor{lightblue} 
        OpenThinker2-7B & Qwen2.5-7B-Instruct & 7B & SFT \\
        \rowcolor{lightblue} 
        OpenThinker3-7B & Qwen2.5-7B-Instruct & 7B & SFT \\
        \rowcolor{lightpink} 
        Qwen-2.5-1.5B-SimpleRL-Zoo & Qwen2.5-1.5B-Base & 1.5B & RL \\
        \rowcolor{lightpink} 
        Qwen-2.5-7B-SimpleRL-Zoo & Qwen2.5-7B-Base & 7B & RL \\
        \rowcolor{lightpink} 
        Qwen-2.5-14B-SimpleRL-Zoo & Qwen2.5-14B-Base & 14B & RL \\
        \rowcolor{lightpink} 
        Qwen-2.5-Math-7B-SimpleRL-Zoo & Qwen2.5-Math-7B-Base & 7B & RL \\
        \rowcolor{lightpink} 
        Qwen2.5-Math-7B-Dr.GRPO & Qwen2.5-Math-7B-Base & 7B & RL \\
        \rowcolor{lightblue} 
        s1.1-7B & Qwen2.5-7B-Instruct & 7B & SFT \\
        \rowcolor{lightpink} 
        DAPO-Qwen-32B & Qwen2.5-32B-Base & 32B & RL \\
        \rowcolor{lightblue} 
        OpenThinker2-32B & Qwen2.5-32B-Instruct & 32B & SFT \\
        \rowcolor{lightblue} 
        s1.1-32B & Qwen2.5-32B-Instruct & 32B & SFT \\
        \bottomrule[1.2pt]
    \end{tabular}
    \caption{\textbf{The Overview of the Open-source LLMs Used in Observational Study}, including their initial model, parameter size, and training paradigm.}
    \label{tab:appendix1_open_source_model}
\end{table*}

%% file: tabs/appendix_12_robustness_of_MTI.tex
\begin{table}[!h]
\centering
\small
\renewcommand{\arraystretch}{1.15}
\setlength{\tabcolsep}{10pt}
\begin{tabular}{lccc}
\toprule
\textbf{Setting} & \textbf{MTI} & \textbf{NGR} & \textbf{RGR} \\
\midrule
En-only      & 1.163 & 0.396 & 0.732 \\
+1 parallel  & 2.496 & 0.963 & 1.480 \\
+2 parallels & 2.650 & 1.061 & 1.601 \\
\bottomrule
\end{tabular}

\caption{\textbf{Robustness of transfer metrics on the main parallel-training comparison.}
We compare the original MTI with normalized-gain ratio (NGR) and raw-gain
ratio (RGR). All three metrics show a large gain from English-only to
one-parallel training, followed by a smaller additional gain.}
\label{tab:metric_robustness_summary}
\end{table}

%% file: tabs/appendix_11_sensitivity_to_base_acc.tex
\begin{table}[t]
\centering
\small
\renewcommand\arraystretch{1.0}
\setlength{\tabcolsep}{1.2mm}
\begin{tabular}{lccc}
\toprule[1.2pt]
\textbf{Setting} & \textbf{All} & $\mathbf{Acc^{base}>20}$ & $\mathbf{Acc^{base}>30}$ \\
\midrule[0.8pt]
\multicolumn{4}{c}{\textit{Relative-gain MTI}} \\
En-only     & 1.16 & 1.16 & 1.23 \\
+1 parallel & 2.50 & 2.42 & 1.88 \\
+2 parallels& 2.65 & 2.62 & 2.13 \\
+3 parallels& 3.00 & 3.16 & 2.58 \\
+4 parallels& 3.28 & 3.25 & 2.34 \\
+5 parallels& 3.48 & 3.41 & 2.02 \\
\midrule[0.8pt]
\multicolumn{4}{c}{\textit{Normalized-gain ratio}} \\
En-only     & 0.40 & 0.40 & 0.48 \\
+1 parallel & 0.96 & 0.95 & 1.05 \\
+2 parallels& 1.06 & 1.08 & 1.22 \\
+3 parallels& 1.15 & 1.19 & 1.37 \\
+4 parallels& 1.18 & 1.20 & 1.33 \\
+5 parallels& 1.63 & 1.57 & 1.85 \\
\midrule[0.8pt]
\multicolumn{4}{c}{\textit{Raw-gain ratio}} \\
En-only     & 0.84 & 0.84 & 0.85 \\
+1 parallel & 1.45 & 1.47 & 1.46 \\
+2 parallels& 1.61 & 1.67 & 1.68 \\
+3 parallels& 1.95 & 2.04 & 1.97 \\
+4 parallels& 1.85 & 1.88 & 1.81 \\
+5 parallels& 2.06 & 2.11 & 2.08 \\
\bottomrule[1.2pt]
\end{tabular}
\caption{\textbf{Sensitivity of transfer scores to low-base-accuracy cases.}
Scores are recomputed on multilingual MATH500 after excluding unseen-language
cases with base accuracy below the specified thresholds.}
\label{tab:mti_base_sensitivity}
\end{table}

%% file: tabs/appendix_4_base_model_template.tex
\begin{table*}[!h]
\centering
\footnotesize
\renewcommand\arraystretch{1.0}
\setlength{\tabcolsep}{1.2mm}

\begin{tabular}{lccccccccccc>{\columncolor{lightgray}}c>{\columncolor{lightgray}}c}
\toprule[1.2pt] 
\multirow{2}{*}{\textbf{Settings}} & \multicolumn{11}{c}{\textbf{Accuracy per language}} & \multicolumn{2}{c}{\cellcolor{lightgray}\textbf{Average}} \\ 
\cline{2-12}
& en & es & ru & de & fr & bn & th & sw & zh & ja & te & \textbf{Acc} & \textbf{Off-target} \\
\midrule[0.8pt]
\rowcolor{lightblue} 
\multicolumn{14}{c}{\textit{Qwen2.5-7B-Base}} \\

Qwen-Math Template & 49.2 & 31.8 & 25.2 & 28.2 & 30.0 & 5.8 & 23.0 & 2.4 & 27.0 & 15.2 & 1.4 & 21.7 & 16.6 \\
Qwen-Instruct Template & 50.6 & 38.0 & 30.0 & 33.2 & 38.4 & 10.4 & 26.8 & 2.4 & 30.0 & 27.8 & 4.4 & 26.5 & 15.7 \\
No Template & 44.4 & 38.2 & 28.2 & 28.4 & 35.6 & 6.0 & 17.0 & 0.2 & 29.2 & 19.8 & 1.2 & 22.6 & 18.5 \\
\midrule[0.8pt]
\rowcolor{lightblue} 
\multicolumn{14}{c}{\textit{Qwen2.5-Math-7B-Base}} \\

Qwen-Math Template & 43.4 & 36.6 & 2.8 & 21.8 & 21.4 & 26.2 & 15.0 & 2.2 & 36.6 & 9.4 & 1.2 & 19.7 & 30.5 \\
Qwen-Instruct Template & 56.6 & 46.4 & 11.2 & 33.4 & 36.8 & 31.4 & 28.2 & 4.2 & 44.4 & 25.2 & 3.2 & 29.2 & 18.0 \\
No Template & 37.8 & 33.0 & 4.2 & 29.8 & 37.4 & 29.0 & 12.4 & 0.0 & 36.2 & 17.2 & 3.0 & 21.8 & 31.5 \\
\bottomrule[1.2pt]
\end{tabular}
\caption{\textbf{The Performance of Base Models with Different Template Settings.} Accuracy (\%) and Off-target rate (\%) across languages for different template settings on multilingual MATH500 benchmark.}
\label{tab:appendix4_base_model_template}
\end{table*}

%% file: tabs/appendix_5_observational_study_initial_model_acc_and_off-tag.tex
\begin{table*}[!h]
\centering
\footnotesize
\renewcommand\arraystretch{1.0}
\setlength{\tabcolsep}{1.0mm}

\begin{tabular}{lccccccccccc>{\columncolor{lightgray}}c>{\columncolor{lightgray}}c}
\toprule[1.2pt] 
\multirow{2}{*}{\textbf{Settings}} & \multicolumn{11}{c}{\textbf{Accuracy per language}} & \multicolumn{2}{c}{\cellcolor{lightgray}\textbf{Average}} \\ 
\cline{2-12}
& en & es & ru & de & fr & bn & th & sw & zh & ja & te & \textbf{Acc} & \textbf{Off-target} \\
\midrule[0.8pt]
\rowcolor{lightblue} 
\multicolumn{14}{c}{\textit{Multilingual MATH500}} \\
Qwen2.5-1.5B & 19.60 & 10.60 & 7.80 & 1.00 & 9.80 & 1.00 & 3.00 & 0.00 & 8.60 & 2.20 & 0.00 & 5.78 & 21.02 \\
Qwen2.5-Math-7B & 56.60 & 46.40 & 11.20 & 33.40 & 36.80 & 31.40 & 28.20 & 4.20 & 44.40 & 25.20 & 3.20 & 29.18 & 17.96 \\
Qwen2.5-7B-Instruct & 74.80 & 69.00 & 59.60 & 56.60 & 62.60 & 37.60 & 49.80 & 18.00 & 53.00 & 52.40 & 26.60 & 50.91 & 0.18 \\
Qwen2.5-7B & 50.60 & 38.00 & 30.00 & 33.20 & 38.40 & 10.40 & 26.80 & 2.40 & 30.00 & 27.80 & 4.40 & 26.55 & 15.69 \\
Qwen2.5-14B & 42.20 & 40.40 & 36.00 & 29.60 & 36.20 & 22.60 & 26.60 & 5.60 & 18.80 & 25.20 & 5.60 & 26.25 & 15.55 \\
Qwen2.5-32B & 54.00 & 50.80 & 42.00 & 37.80 & 46.80 & 24.60 & 33.00 & 13.60 & 28.00 & 42.60 & 11.60 & 34.98 & 5.56 \\
Qwen2.5-32B-Instruct & 78.60 & 73.60 & 68.00 & 68.40 & 69.40 & 53.20 & 60.60 & 37.40 & 61.40 & 65.00 & 43.40 & 61.73 & 0.13 \\
\midrule[0.8pt]
\rowcolor{lightblue} 
\multicolumn{14}{c}{\textit{Multilingual AIME24}} \\
Qwen2.5-1.5B & 0.21 & 0.42 & 0.00 & 0.00 & 0.00 & 0.00 & 0.21 & 0.00 & 0.63 & 0.00 & 0.00 & 0.13 & 19.26 \\
Qwen2.5-Math-7B & 13.75 & 6.46 & 1.67 & 3.33 & 4.79 & 2.71 & 3.33 & 0.00 & 6.88 & 1.67 & 0.63 & 4.11 & 21.76 \\
Qwen2.5-7B-Instruct & 10.42 & 8.96 & 8.13 & 8.75 & 8.54 & 2.92 & 4.58 & 1.04 & 5.63 & 4.38 & 1.88 & 5.93 & 0.34 \\
Qwen2.5-7B & 2.29 & 1.67 & 2.08 & 2.71 & 2.08 & 0.21 & 0.63 & 0.00 & 1.25 & 0.63 & 0.00 & 1.23 & 9.89 \\
Qwen2.5-14B & 2.50 & 2.50 & 2.29 & 2.50 & 2.71 & 0.63 & 0.21 & 0.00 & 1.04 & 0.83 & 0.21 & 1.40 & 16.02 \\
Qwen2.5-32B & 2.71 & 3.13 & 2.08 & 3.33 & 2.71 & 0.21 & 1.04 & 0.00 & 1.25 & 2.08 & 0.00 & 1.69 & 4.07 \\
Qwen2.5-32B-Instruct & 15.63 & 12.71 & 11.25 & 11.67 & 12.08 & 5.42 & 7.50 & 2.92 & 7.29 & 10.63 & 2.50 & 9.05 & 0.51 \\
\midrule[0.8pt]
\rowcolor{lightblue} 
\multicolumn{14}{c}{\textit{Multilingual AIME25}} \\
Qwen2.5-1.5B & 0.00 & 0.00 & 0.00 & 0.00 & 0.00 & 0.00 & 0.00 & 0.00 & 0.21 & 0.00 & 0.00 & 0.02 & 61.14 \\
Qwen2.5-Math-7B & 6.04 & 3.13 & 0.83 & 1.46 & 2.29 & 0.42 & 0.83 & 0.00 & 4.79 & 0.42 & 0.42 & 1.88 & 23.47 \\
Qwen2.5-7B-Instruct & 7.08 & 5.63 & 5.21 & 3.96 & 4.79 & 0.83 & 1.67 & 0.00 & 3.75 & 2.71 & 0.42 & 3.28 & 0.55 \\
Qwen2.5-7B & 0.83 & 0.83 & 0.21 & 1.25 & 0.63 & 0.21 & 0.00 & 0.00 & 0.42 & 0.21 & 0.00 & 0.42 & 10.38 \\
Qwen2.5-14B & 1.25 & 2.08 & 2.50 & 1.67 & 1.46 & 0.00 & 0.21 & 0.00 & 0.42 & 1.46 & 0.21 & 1.02 & 16.99 \\
Qwen2.5-32B & 1.04 & 1.04 & 1.46 & 0.21 & 0.83 & 0.21 & 0.00 & 0.00 & 1.04 & 1.04 & 0.21 & 0.64 & 4.02 \\
Qwen2.5-32B-Instruct & 11.25 & 7.29 & 7.29 & 6.04 & 6.67 & 1.25 & 2.71 & 0.00 & 5.42 & 2.71 & 0.42 & 4.64 & 0.30 \\

\midrule[0.8pt]
\rowcolor{lightblue} 
\multicolumn{14}{c}{\textit{Multilingual GPQA-Diamond}} \\
Qwen2.5-1.5B & 15.66 & 14.65 & 15.15 & 16.67 & 11.62 & 7.58 & 15.15 & 13.64 & 15.15 & 6.06 & 15.15 & 13.31 & 20.02 \\
Qwen2.5-Math-7B & 16.16 & 13.64 & 3.54 & 17.17 & 15.66 & 17.68 & 17.68 & 11.62 & 22.22 & 0.51 & 16.16 & 13.82 & 27.18 \\
Qwen2.5-7B-Instruct & 36.36 & 32.83 & 23.74 & 36.36 & 33.84 & 26.77 & 29.80 & 24.75 & 30.81 & 29.80 & 21.72 & 29.71 & 0.64 \\
Qwen2.5-7B & 28.79 & 24.24 & 20.71 & 22.22 & 18.18 & 11.11 & 21.72 & 17.68 & 23.23 & 17.17 & 12.63 & 19.79 & 9.69 \\
Qwen2.5-14B & 26.26 & 15.15 & 20.71 & 21.21 & 27.78 & 20.20 & 24.24 & 22.22 & 12.12 & 10.61 & 16.16 & 19.70 & 20.98 \\
Qwen2.5-32B & 28.28 & 30.30 & 28.28 & 33.33 & 23.74 & 15.66 & 24.75 & 17.68 & 27.27 & 27.78 & 17.17 & 24.93 & 9.00 \\
Qwen2.5-32B-Instruct & 45.45 & 41.92 & 38.89 & 41.41 & 44.44 & 29.80 & 38.38 & 32.83 & 36.36 & 38.38 & 27.27 & 37.74 & 0.28 \\

\bottomrule[1.2pt]
\end{tabular}
\caption{\textbf{The Performance of Initial Models.} Accuracy (\%) and Off-target rate (\%) across languages for different Initial models.}
\label{tab:appendix5_initial_model_acc_and_off-tag}
\end{table*}

%% file: tabs/appendix_10_the_impact_of_hyperparameter_lambda.tex
\begin{table*}[!h]
\centering
\small
\setlength{\tabcolsep}{3pt}
\begin{tabular}{l c ccccccccccc ccc}
\toprule[1.2pt]
\multirow{2}{*}{\textbf{Setting}} &
\multirow{2}{*}{$(\lambda_1,\lambda_2,\lambda_3)$} &
\multicolumn{11}{c}{\textbf{Accuracy per Language}} &
\multirow{2}{*}{\textbf{Avg.}} &
\multirow{2}{*}{\textbf{Off-tgt.}} &
\multirow{2}{*}{\textbf{MTI}} \\
\cmidrule(lr){3-13}
& & \textbf{en} & \textbf{es} & \textbf{ru} & \textbf{de} & \textbf{fr}
& \textbf{bn} & \textbf{th} & \textbf{sw} & \textbf{zh} & \textbf{ja}
& \textbf{te} & & & \\
\midrule[0.8pt]
\rowcolor{lightblue}
C1 (Balanced)
& (0.8, 0.1, 0.1)
& 78.4 & 73.4 & 66.0 & 65.6 & 67.2 & 48.8 & 57.4 & 26.2
& 57.8 & 62.0 & 33.8 & \textbf{57.87} & \textbf{0.20} & \textbf{2.50} \\

C2 (w/o LCR)
& (0.9, 0.1, 0.0)
& 77.8 & 71.4 & 66.2 & 63.4 & 68.0 & 44.2 & 56.4 & 26.2
& 59.6 & 57.8 & 31.8 & 56.62 & 1.24 & 2.10 \\

C3 (Strong LCR)
& (0.7, 0.1, 0.2)
& 77.2 & 70.2 & 65.8 & 65.8 & 67.2 & 43.6 & 55.4 & 25.6
& 60.2 & 56.2 & 32.6 & 56.35 & 0.38 & 2.25 \\

C4 (w/o FR)
& (0.9, 0.0, 0.1)
& 75.2 & 60.8 & 62.2 & 56.0 & 60.2 & 36.2 & 43.8 & 18.0
& 54.4 & 31.8 & 24.8 & 47.58 & 0.31 & -3.45 \\

C5 (Strong FR)
& (0.7, 0.2, 0.1)
& 78.2 & 71.6 & 65.2 & 62.6 & 67.8 & 45.0 & 60.0 & 24.8
& 60.0 & 60.8 & 29.2 & 56.84 & 0.35 & 2.23 \\
\bottomrule[1.2pt]
\end{tabular}

\caption{\textbf{Sensitivity analysis of reward hyperparameters.}
Models are trained on English data with parallel Russian data.
$(\lambda_1,\lambda_2,\lambda_3)$ denotes the weights of the three reward
components. 
% Avg. and Off-tgt. denote average accuracy and average off-target language rate, respectively. MTI denotes the Multilingual Transfer Index.
}
\label{tab:appendix_10_the_impact_of_hyperparameter_lambda}
\end{table*}

%% file: tabs/experiment_2_control_study_initial_model.tex
\begin{table}[!h]
\centering
\small
\setlength{\tabcolsep}{2.2pt}
\renewcommand{\arraystretch}{1.1}
\begin{tabular}{lrrrrrrr}
\toprule
\textbf{Setting}
& \textbf{M500}
& \textbf{A24}
& \textbf{A25}
& \textbf{GPQA}
& \textbf{Avg.}
& \textbf{Off-tgt.}
& \textbf{MTI} \\
\midrule

\rowcolor{lightblue}
Base
& 26.55 & 1.23 & 0.42 & 19.79 & 12.00 & 11.41 & -- \\
\hspace{4pt}$\looparrowright$ GRPO
& 52.16 & 7.10 & 3.35 & 27.18 & 22.45 & 3.12 & 1.95 \\

\rowcolor{lightblue}
Instruct
& 50.91 & 5.93 & 3.28 & 29.71 & 22.45 & 1.43 & -- \\
\hspace{4pt}$\looparrowright$ GRPO
& 54.24 & 7.41 & 3.92 & 28.47 & 23.51 & 0.94 & 1.23 \\

\rowcolor{lightblue}
Math
& 29.18 & 4.11 & 1.88 & 13.82 & 12.25 & 22.59 & -- \\
\hspace{4pt}$\looparrowright$ GRPO
& 45.73 & 8.84 & 3.96 & 18.96 & 19.37 & 9.50 & 2.12 \\

\bottomrule
\end{tabular}

\caption{\textbf{Impact of model initialization on cross-lingual transfer.}
All models are initialized from Qwen2.5-7B variants and trained with GRPO
on the same English data. We report average accuracy (\%) across languages
on MATH500 (M500), AIME24 (A24), AIME25 (A25), and GPQA-Diamond (GPQA),
together with overall average accuracy, off-target rate (\%), and MTI.}
\label{tab:experiment_2_control_study_initial_model}
\end{table}

%% file: tabs/appendix_13_sino_tibetan_family.tex
\begin{table}[!h]
\centering
\small
\renewcommand\arraystretch{1.0}
\setlength{\tabcolsep}{4pt}
\begin{tabular}{lccc}
\toprule[1.2pt]
\textbf{Model} & \textbf{bo} & \textbf{my} & \textbf{zh} \\
\midrule[0.8pt]

\textit{Qwen2.5-7B-Instruct}
& 10.40 & 9.20 & 53.00 \\
\quad + GRPO (En)
& 13.60 & 11.60 & 58.60 \\
\quad $\Delta$
& (+3.20) & (+2.40) & (+5.60) \\

\midrule[0.8pt]

\textit{Llama-3.1-8B-Instruct}
& 0.80 & 2.40 & 28.40 \\
\quad + GRPO (En)
& 15.00 & 8.40 & 33.80 \\
\quad $\Delta$
& \textbf{(+14.20)} & \textbf{(+6.00)} & \textbf{(+5.40)} \\

\bottomrule[1.2pt]
\end{tabular}

\caption{\textbf{Cross-lingual transfer to Sino-Tibetan languages.}
We report accuracy (\%) before and after English-only RPT with GRPO.
\emph{bo}, \emph{my}, and \emph{zh} denote Tibetan, Myanmar, and Chinese,
respectively. $\Delta$ denotes the absolute accuracy gain after GRPO.}
\label{tab:sino_tibetan}
\end{table}

%% file: tabs/appendix_8_control_study_scaling_up.tex
\begin{table*}[!h]
\centering
\small
\renewcommand\arraystretch{1.0}
\setlength{\tabcolsep}{1.0mm}

\begin{tabular}{lccccccccccc>{\columncolor{lightgray}}c>{\columncolor{lightgray}}c}
\toprule[1.2pt] 
\multirow{2}{*}{\textbf{Settings}} & \multicolumn{11}{c}{\textbf{$\Delta$ Performance}} & \multicolumn{2}{c}{\cellcolor{lightgray}\textbf{Average across languages}} \\ 
\cline{2-12}
& en & es & ru & de & fr & bn & th & sw & zh & ja & te & \textit{Training} & \textit{Untraining} \\
\midrule[0.8pt]
\rowcolor{lightblue} 
\multicolumn{14}{c}{\textit{Qwen2.5-1.5B-Instruct with GRPO on En Data}} \\
MATH500 & 20.40 & 27.20 & 17.40 & 14.40 & 17.40 & 6.20 & 7.80 & 4.40 & 16.80 & 14.20 & 7.40 & 20.40 & 13.32 \\
AIME24 & 2.08 & 0.42 & -0.21 & 1.25 & 0.42 & 0.21 & 0.21 & 0.00 & -0.21 & 0.21 & 0.21 & 2.08 & 0.25 \\
AIME25 & 1.04 & 0.21 & 0.00 & 0.21 & 0.00 & 0.00 & 0.00 & -0.21 & 0.42 & 0.21 & 0.21 & 1.04 & 0.10 \\
GPQA-Diamond & 9.09 & 20.71 & -0.51 & 5.05 & 12.12 & 1.52 & -2.53 & 2.02 & 8.59 & 3.54 & 0.51 & 9.09 & 5.10 \\

\midrule[0.8pt]
\rowcolor{lightblue} 
\multicolumn{14}{c}{\textit{Qwen2.5-7B-Instruct with GRPO on En Data}} \\
MATH500 & 4.40 & 1.00 & 1.40 & 5.60 & 5.60 & 4.20 & 5.60 & 1.40 & 0.40 & 6.20 & 0.80 & 4.40 & 3.22 \\
AIME24 & 2.71 & 1.04 & 0.83 & 1.46 & 0.42 & 0.42 & 0.42 & 2.08 & 2.08 & 4.58 & 0.21 & 2.71 & 1.35 \\
AIME25 & 1.25 & -1.04 & 0.00 & 2.50 & 0.00 & 1.46 & 1.67 & 0.00 & 0.63 & 0.83 & -0.21 & 1.25 & 0.58 \\
GPQA-Diamond & -3.54 & 4.55 & 0.00 & -1.01 & 7.07 & -2.02 & -1.01 & -7.07 & 4.04 & -3.03 & 0.51 & -3.54 & 0.20 \\

\bottomrule[1.2pt]
\end{tabular}
\caption{\textbf{The Impact of Model Size in Interventional Study.} $\Delta$ Performance on various benchmarks across \textit{Qwen2.5-1.5B-Instruct} and \textit{Qwen2.5-7B-Instruct}.}
\label{tab:appendix8_control_strudy_scaling_up}
\end{table*}

%% file: tabs/appendix_2_scaling_parallel_law_acc.tex
\begin{table*}[!h]
\centering
\small
\renewcommand\arraystretch{1.0}
\setlength{\tabcolsep}{1.2mm}

\begin{tabular}{lccccccccccc>{\columncolor{lightgray}}c>{\columncolor{lightgray}}c>{\columncolor{lightgray}}c}
\toprule[1.2pt] 
\multirow{2}{*}{\textbf{Settings}} & \multicolumn{11}{c}{\textbf{Accuracy per language}} & \multicolumn{3}{c}{\cellcolor{lightgray}\textbf{Average}} \\ 
\cline{2-12}
& en & es & ru & de & fr & bn & th & sw & zh & ja & te & \textbf{Acc} & \textbf{Off-target} & \textbf{MTI} \\
\midrule[0.8pt]
\rowcolor{lightblue} 
\multicolumn{15}{c}{\textit{Multilingual MATH500}} \\
Only English & 79.2 & 70.0 & 61.0 & 62.2 & 68.2 & 41.8 & 55.4 & 19.4 & 53.4 & 58.6 & 27.4 & 54.2 & 0.5 & 1.163 \\
\hspace{4pt} \emph{w.} One parallel & 78.4 & 73.4 & 66.0 & 65.6 & 67.2 & 48.8 & 57.4 & 26.2 & 57.8 & 62.0 & 33.8 & 57.9 & 0.2 & 2.496 \\
\hspace{4pt} \emph{w.} Two parallel & 79.0 & 73.4 & 64.4 & 67.4 & 69.2 & 45.2 & 60.2 & 26.2 & 63.0 & 61.6 & 32.6 & 58.4 & 0.2 & 2.650 \\
\hspace{4pt} \emph{w.} Three parallel & 77.8 & 73.6 & 64.6 & 68.4 & 69.8 & 46.0 & 60.8 & 24.0 & 62.2 & 60.8 & 34.2 & 58.4 & 0.4 & 3.002 \\
\hspace{4pt} \emph{w.} Four parallel & 77.2 & 71.2 & 66.2 & 66.8 & 68.0 & 47.6 & 61.8 & 28.6 & 62.0 & 60.2 & 35.2 & 58.6 & 0.6  & 3.282 \\
\hspace{4pt} \emph{w.} Five parallel & 77.4 & 71.4 & 62.2 & 66.2 & 66.0 & 48.6 & 62.0 & 32.4 & 62.2 & 63.8 & 37.0 & 59.0 & 0.4 & 3.475 \\
\hspace{4pt} \emph{w.} Six parallel & 76.4 & 70.8 & 63.8 & 65.6 & 66.8 & 48.6 & 61.8 & 34.6 & 63.4 & 63.4 & 38.4 & 59.4 & 0.5 & 3.534 \\
\hspace{4pt} \emph{w.} Seven parallel & 76.6 & 71.2 & 63.6 & 66.2 & 66.2 & 49.4 & 62.6 & 33.8 & 63.5 & 63.4 & 38.2 & 59.5 & 0.2 & 3.631 \\
\bottomrule[1.2pt]
\end{tabular}
\caption{The Detailed Results in Parallel Scaling Law.}
\label{tab:appendix2_scaling_parallel_law_acc}
\end{table*}

%% file: tabs/appendix_9_scaling_parallel_law_languages.tex
\begin{table}[!h]
\centering
\footnotesize
\renewcommand\arraystretch{1.0}
\setlength{\tabcolsep}{1.2mm}

\begin{tabular}{ll}
\toprule[1.2pt] 
Settings& Training Parallel Languages \\
\midrule[0.8pt]
% \rowcolor{lightblue} 
Only English & \textit{en} \\
\hspace{4pt} \emph{w.} One parallel & \textit{en, ru} \\
\hspace{4pt} \emph{w.} Two parallel & \textit{en, ru, fr} \\
\hspace{4pt} \emph{w.} Three parallel & \textit{en, ru, fr, es} \\
\hspace{4pt} \emph{w.} Four parallel & \textit{en, ru, fr, es, de} \\
\hspace{4pt} \emph{w.} Five parallel & \textit{en, ru, fr, es, de, bn} \\
\hspace{4pt} \emph{w.} Six parallel & \textit{en, ru, fr, es, de, bn, th} \\
\hspace{4pt} \emph{w.} Seven parallel & \textit{en, ru, fr, es, de, bn, th, zh} \\

\bottomrule[1.2pt]
\end{tabular}
\caption{The Language Settings in Parallel Scaling Law.}
\label{tab:appendix9_scaling_parallel_law_languages}
\end{table}

% \begin{wraptable}{r}{0.45\textwidth} % r表示右边，0.45\textwidth表示表格宽度
% \centering
% \footnotesize
% \renewcommand\arraystretch{1.2}
% \setlength{\tabcolsep}{1.2mm}
% \caption{\textbf{The Language Settings in Parallel Scaling Law.}}
% \label{tab:appendix9_scaling_parallel_law_languages}
% \begin{tabular}{ll}
% \toprule[1.2pt] 
% Settings & Training Parallel Languages \\
% \midrule[0.8pt]
% Only English & en \\
% \hspace{4pt} \emph{w.} One parallel & en, ru \\
% \hspace{4pt} \emph{w.} Two parallel & en, ru, fr \\
% \hspace{4pt} \emph{w.} Three parallel & en, ru, fr, es \\
% \hspace{4pt} \emph{w.} Four parallel & en, ru, fr, es, de \\
% \hspace{4pt} \emph{w.} Five parallel & en, ru, fr, es, de, bn \\
% \hspace{4pt} \emph{w.} Six parallel & en, ru, fr, es, de, bn, th \\
% \hspace{4pt} \emph{w.} Seven parallel & en, ru, fr, es, de, bn, th, zh \\
% \bottomrule[1.2pt]
% \end{tabular}
% \end{wraptable}

%% file: tabs/appendix_12_chinese-centric.tex
\begin{table}[!h]
\centering
\small
\renewcommand{\arraystretch}{1.2}
\setlength{\tabcolsep}{5pt}
\begin{tabular}{lrrrr}
\toprule[1.2pt]
\textbf{Lang.}
& \textbf{Base}
& \textbf{Zh}
& \textbf{Zh+Ru}
& \textbf{Zh+Ru+Fr} \\
\midrule[0.8pt]
en & 74.8 & 76.8 & 78.8 & 78.0 \\
es & 69.0 & 70.4 & 72.0 & 72.2 \\
ru & 59.6 & 61.8 & 64.4 & 63.6 \\
de & 56.6 & 62.2 & 65.0 & 66.4 \\
fr & 62.6 & 66.8 & 70.0 & 69.8 \\
bn & 37.6 & 43.0 & 44.4 & 45.4 \\
th & 49.8 & 55.2 & 60.0 & 61.0 \\
sw & 18.0 & 21.8 & 28.6 & 30.0 \\
zh & 53.0 & 59.4 & 61.4 & 62.8 \\
ja & 52.4 & 56.6 & 60.4 & 64.0 \\
te & 26.6 & 31.0 & 33.4 & 33.2 \\
\midrule[0.8pt]
Avg.     & 50.91 & 55.00 & 58.13 & \textbf{58.76} \\
Off-tgt. & 1.18  & 0.38  & 0.26  & \textbf{0.23} \\
MTI      & --    & 0.86  & 1.69  & \textbf{1.87} \\
\bottomrule[1.2pt]
\end{tabular}

\caption{\textbf{Parallel scaling with Chinese as the source language.}
All experiments use Qwen2.5-7B-Instruct and are evaluated on multilingual
MATH500. Adding the first parallel language yields the largest gain, while
adding a second language provides a smaller additional improvement.}
\label{tab:chinese_parallel_scaling}
\end{table}

%% file: tabs/appendix_7_observational_study_open_models_acc_and_off-tag.tex
\begin{table*}[htbp]
\centering
\footnotesize
\renewcommand\arraystretch{1.0}
\setlength{\tabcolsep}{0.5mm}

\begin{tabular}{lccccccccccc>{\columncolor{lightgray}}c>{\columncolor{lightgray}}c}
\toprule[1.2pt] 
\multirow{2}{*}{\textbf{Settings}} & \multicolumn{11}{c}{\textbf{Accuracy per language}} & \multicolumn{2}{c}{\cellcolor{lightgray}\textbf{Average}} \\ 
\cline{2-12}
& en & es & ru & de & fr & bn & th & sw & zh & ja & te & \textbf{Acc} & \textbf{Off-target} \\
\midrule[0.8pt]
\rowcolor{lightblue} 
\multicolumn{14}{c}{\textit{Multilingual MATH500}} \\
DeepSeek-R1-Distill-Qwen-7B & 86.20 & 76.20 & 67.00 & 66.40 & 69.80 & 40.20 & 53.80 & 18.40 & 70.00 & 53.20 & 17.60 & 56.25 & 2.69 \\
Open-Reasoner-Zero-7B & 81.60 & 76.00 & 70.40 & 69.80 & 71.20 & 45.20 & 63.20 & 15.00 & 62.80 & 61.80 & 17.60 & 57.69 & 5.20 \\
OpenThinker2-7B & 86.00 & 74.80 & 64.80 & 63.00 & 73.00 & 34.40 & 58.80 & 12.60 & 65.80 & 70.00 & 8.40 & 55.60 & 36.13 \\
OpenThinker3-7B & 85.80 & 81.80 & 75.00 & 69.20 & 76.40 & 17.00 & 48.80 & 17.40 & 69.60 & 65.00 & 10.40 & 56.04 & 60.75 \\
Qwen-2.5-1.5B-SimpleRL-Zoo & 57.60 & 41.40 & 36.80 & 38.20 & 42.40 & 11.80 & 28.80 & 14.60 & 33.60 & 31.00 & 7.40 & 31.24 & 21.51 \\
Qwen-2.5-7B-SimpleRL-Zoo & 77.60 & 72.00 & 65.00 & 64.20 & 68.00 & 42.20 & 60.40 & 24.20 & 62.00 & 59.80 & 25.80 & 56.47 & 4.31 \\
Qwen-2.5-14B-SimpleRL-Zoo & 82.40 & 74.40 & 71.00 & 69.40 & 73.60 & 54.80 & 69.00 & 33.20 & 65.60 & 68.80 & 41.00 & 63.93 & 0.47 \\
Qwen-2.5-Math-7B-SimpleRL-Zoo & 80.40 & 72.60 & 66.00 & 68.60 & 70.60 & 45.80 & 57.40 & 15.00 & 61.80 & 54.40 & 14.20 & 55.16 & 8.33 \\
Qwen2.5-Math-7B-Oat-Zero & 79.80 & 72.40 & 32.80 & 55.60 & 50.40 & 47.60 & 49.40 & 16.80 & 58.00 & 43.40 & 11.80 & 47.09 & 15.11 \\
s1.1-7B & 75.80 & 68.20 & 60.80 & 59.00 & 69.60 & 37.00 & 57.40 & 16.40 & 57.20 & 51.60 & 20.40 & 52.13 & 12.76 \\
DAPO-Qwen-32B & 68.80 & 65.00 & 58.80 & 60.20 & 63.00 & 52.80 & 58.40 & 44.80 & 54.20 & 56.80 & 44.80 & 57.05 & 11.85 \\
OpenThinker2-32B & 96.00 & 88.60 & 85.20 & 84.20 & 85.00 & 73.00 & 80.60 & 35.40 & 77.40 & 75.60 & 47.20 & 75.29 & 13.02 \\
S1.1-32B & 95.40 & 91.20 & 85.00 & 83.60 & 88.00 & 73.00 & 81.20 & 57.00 & 77.40 & 80.80 & 53.60 & 78.75 & 27.40 \\

\midrule[0.8pt]
\rowcolor{lightblue} 
\multicolumn{14}{c}{\textit{Multilingual AIME24}} \\
DeepSeek-R1-Distill-Qwen-7B & 40.63 & 27.71 & 26.25 & 23.13 & 27.50 & 6.46 & 9.79 & 2.50 & 30.42 & 9.79 & 0.83 & 18.64 & 7.77 \\
Open-Reasoner-Zero-7B & 16.25 & 18.13 & 17.29 & 15.21 & 17.71 & 9.58 & 14.79 & 1.67 & 14.79 & 14.79 & 1.25 & 12.86 & 10.78 \\
OpenThinker2-7B & 37.08 & 18.33 & 17.08 & 13.75 & 20.83 & 11.04 & 20.21 & 2.50 & 25.63 & 27.29 & 5.42 & 18.11 & 39.72 \\
OpenThinker3-7B & 26.25 & 32.08 & 23.54 & 26.46 & 29.38 & 5.63 & 16.25 & 3.54 & 26.46 & 19.38 & 3.54 & 19.32 & 63.28 \\
Qwen-2.5-1.5B-SimpleRL-Zoo & 0.00 & 0.00 & 0.42 & 0.00 & 0.21 & 0.00 & 0.21 & 0.42 & 1.25 & 0.21 & 0.00 & 0.25 & 60.42 \\
Qwen-2.5-7B-SimpleRL-Zoo & 6.25 & 7.29 & 6.25 & 7.29 & 8.33 & 3.75 & 5.42 & 2.08 & 5.42 & 4.38 & 2.50 & 5.36 & 77.16 \\
Qwen-2.5-14B-SimpleRL-Zoo & 12.71 & 13.13 & 13.13 & 10.42 & 13.33 & 9.17 & 9.79 & 3.54 & 10.42 & 11.46 & 5.63 & 10.25 & 0.42 \\
Qwen-2.5-Math-7B-SimpleRL-Zoo & 25.83 & 15.42 & 13.96 & 11.25 & 13.33 & 5.00 & 8.13 & 1.67 & 10.21 & 8.54 & 2.50 & 10.53 & 11.29 \\
Qwen2.5-Math-7B-Oat-Zero & 28.33 & 12.92 & 5.42 & 8.54 & 11.67 & 6.25 & 8.75 & 1.04 & 12.29 & 6.67 & 0.42 & 9.30 & 20.57 \\
s1.1-7B & 14.38 & 10.42 & 10.42 & 10.21 & 11.46 & 5.21 & 7.92 & 1.67 & 8.75 & 7.71 & 0.21 & 8.03 & 7.95 \\
DAPO-Qwen-32B & 54.58 & 50.00 & 51.67 & 46.04 & 50.00 & 42.50 & 36.04 & 19.17 & 40.83 & 45.42 & 27.29 & 42.14 & 5.91 \\
OpenThinker2-32B & 74.17 & 61.88 & 55.42 & 56.67 & 55.42 & 59.17 & 49.17 & 13.96 & 56.88 & 37.71 & 34.58 & 50.45 & 22.08 \\
S1.1-32B & 58.75 & 55.21 & 49.17 & 51.25 & 53.33 & 36.04 & 41.25 & 19.38 & 44.58 & 46.88 & 17.08 & 42.99 & 7.80 \\

\midrule[0.8pt]
\rowcolor{lightblue} 
\multicolumn{14}{c}{\textit{Multilingual AIME25}} \\
DeepSeek-R1-Distill-Qwen-7B & 29.58 & 20.21 & 21.25 & 22.29 & 19.79 & 5.63 & 8.75 & 0.42 & 26.67 & 10.00 & 0.00 & 14.96 & 6.97 \\
Open-Reasoner-Zero-7B & 14.58 & 13.33 & 11.88 & 9.58 & 11.04 & 1.67 & 9.38 & 0.00 & 10.21 & 9.79 & 0.21 & 8.33 & 10.04 \\
OpenThinker2-7B & 28.33 & 21.67 & 20.63 & 17.08 & 21.46 & 9.38 & 17.08 & 2.71 & 25.00 & 24.38 & 2.08 & 17.25 & 39.77 \\
OpenThinker3-7B & 22.50 & 27.71 & 23.33 & 20.00 & 27.50 & 6.67 & 14.79 & 3.13 & 28.54 & 21.67 & 1.67 & 17.95 & 63.28 \\
Qwen-2.5-1.5B-SimpleRL-Zoo & 0.00 & 0.00 & 0.00 & 0.00 & 0.00 & 0.00 & 0.00 & 0.00 & 0.21 & 0.00 & 0.00 & 0.02 & 61.14 \\
Qwen-2.5-7B-SimpleRL-Zoo & 4.58 & 3.33 & 2.08 & 3.96 & 5.42 & 0.83 & 3.13 & 0.63 & 1.46 & 2.50 & 0.21 & 2.56 & 79.51 \\
Qwen-2.5-14B-SimpleRL-Zoo & 13.96 & 11.67 & 10.42 & 10.83 & 10.00 & 3.54 & 6.46 & 1.88 & 6.67 & 8.54 & 2.08 & 7.82 & 0.61 \\
Qwen-2.5-Math-7B-SimpleRL-Zoo & 13.75 & 9.58 & 6.04 & 5.42 & 9.79 & 2.71 & 5.63 & 1.04 & 6.25 & 3.75 & 1.04 & 5.91 & 12.12 \\
Qwen2.5-Math-7B-Oat-Zero & 10.00 & 9.38 & 2.29 & 6.67 & 4.38 & 1.46 & 2.08 & 1.04 & 6.67 & 2.92 & 0.42 & 4.30 & 22.12 \\
s1.1-7B & 13.96 & 11.67 & 9.58 & 6.88 & 11.88 & 2.08 & 7.08 & 0.21 & 9.79 & 5.21 & 0.00 & 7.12 & 6.17 \\
DAPO-Qwen-32B & 38.13 & 38.54 & 37.29 & 36.25 & 34.58 & 30.83 & 32.71 & 18.33 & 31.67 & 34.17 & 22.29 & 32.25 & 4.56 \\
OpenThinker2-32B & 57.29 & 50.00 & 48.13 & 52.29 & 43.96 & 45.42 & 41.88 & 12.50 & 52.50 & 36.04 & 25.63 & 42.33 & 22.65 \\
S1.1-32B & 50.00 & 43.54 & 38.33 & 43.33 & 42.71 & 29.38 & 31.88 & 16.04 & 38.75 & 35.42 & 14.58 & 34.91 & 8.05 \\

\midrule[0.8pt]
\rowcolor{lightblue} 
\multicolumn{14}{c}{\textit{Multilingual GPQA-Diamond}} \\
DeepSeek-R1-Distill-Qwen-7B & 32.32 & 33.33 & 33.33 & 35.35 & 35.35 & 18.18 & 21.21 & 23.74 & 29.80 & 14.65 & 14.14 & 26.49 & 6.11 \\
Open-Reasoner-Zero-7B & 37.37 & 26.77 & 31.82 & 33.33 & 33.33 & 24.24 & 32.83 & 12.63 & 33.33 & 26.77 & 7.07 & 27.23 & 6.20 \\
OpenThinker2-7B & 28.79 & 17.68 & 17.17 & 16.67 & 22.73 & 22.22 & 25.76 & 14.14 & 22.22 & 18.69 & 14.14 & 20.02 & 38.15 \\
OpenThinker3-7B & 23.23 & 18.69 & 22.22 & 16.67 & 24.24 & 5.05 & 14.65 & 12.63 & 21.72 & 10.10 & 7.07 & 16.02 & 59.23 \\
Qwen-2.5-1.5B-SimpleRL-Zoo & 20.71 & 15.66 & 23.74 & 16.67 & 24.75 & 10.10 & 18.18 & 13.13 & 17.17 & 21.21 & 8.59 & 17.26 & 14.69 \\
Qwen-2.5-7B-SimpleRL-Zoo & 30.30 & 31.82 & 29.80 & 31.82 & 33.84 & 20.71 & 23.74 & 17.17 & 29.80 & 23.74 & 10.10 & 25.71 & 3.49 \\
Qwen-2.5-14B-SimpleRL-Zoo & 41.92 & 40.40 & 34.85 & 40.91 & 39.39 & 27.78 & 34.85 & 29.80 & 39.39 & 33.33 & 26.26 & 35.35 & 2.62 \\
Qwen-2.5-Math-7B-SimpleRL-Zoo & 30.81 & 26.26 & 22.73 & 27.78 & 28.28 & 18.18 & 17.17 & 9.09 & 27.27 & 16.67 & 8.08 & 21.12 & 12.26 \\
Qwen2.5-Math-7B-Oat-Zero & 25.76 & 17.17 & 7.07 & 21.21 & 15.66 & 19.19 & 21.72 & 9.09 & 30.81 & 6.06 & 12.63 & 16.94 & 19.74 \\
s1.1-7B & 17.68 & 14.14 & 20.20 & 22.22 & 29.29 & 9.09 & 17.17 & 16.16 & 24.75 & 16.67 & 18.69 & 18.73 & 11.98 \\
DAPO-Qwen-32B & 52.50 & 44.44 & 40.91 & 48.99 & 41.92 & 37.37 & 42.93 & 31.82 & 46.97 & 47.98 & 30.81 & 42.42 & 5.88 \\
OpenThinker2-32B & 62.63 & 57.58 & 58.08 & 59.09 & 58.59 & 50.51 & 47.47 & 21.72 & 56.57 & 0.00 & 0.00 & 42.93 & 22.91 \\
S1.1-32B & 64.65 & 57.58 & 57.58 & 59.60 & 56.57 & 41.41 & 48.48 & 36.36 & 56.57 & 53.03 & 32.83 & 51.33 & 11.85 \\

\bottomrule[1.2pt]
\end{tabular}
\caption{\textbf{The Performance of Various Open-source Models.} Accuracy (\%) and Off-target rate (\%) across languages for various open-source models.}
\label{tab:appendix7_open_models_acc_and_off-tag}
\end{table*}

%% file: prompt/template_for_base_models.tex
\begin{tcolorbox}[colback=lightblue!50!white,colframe=lightblue, width=\textwidth, breakable]
\footnotesize
\label{prompt:template_for_base_model}
% -------------------- Qwen-Instruct Template --------------------
\textbf{Qwen-Instruct Template:}
\begin{verbatim}
<|im_start|>system\n
You are Qwen, created by Alibaba Cloud. You are a helpful assistant.
<|im_end|>\n
<|im_start|>user\n{instruction}<|im_end|>\n
<|im_start|>assistant\n
\end{verbatim}

\textbf{Qwen-Math Template:}
% -------------------- Qwen-Math Template --------------------
\begin{verbatim}
<|im_start|>system\n
Please reason step by step, and put your final answer within \boxed{}.
<|im_end|>\n
<|im_start|>user\n{instruction}<|im_end|>\n
<|im_start|>assistant\n
\end{verbatim}

\textbf{No Template:}

% -------------------- No Template --------------------
\begin{verbatim}
{instruction}
\end{verbatim}

\end{tcolorbox}

%% file: prompt/instruction_used_for_open-source_model.tex
\begin{tcolorbox}[colback=lightblue!50!white,colframe=lightblue,title=\textcolor{black}{The Instruction Used in Multilingual Reasoning Prompt}, width=\textwidth, breakable]
\footnotesize
\label{prompt:multilingual_reasoning_instruction}
Please always think in \texttt{[LANGUAGE]}.\\  
\\
Solve the following mathematics problem step by step. At the end, provide your final answer enclosed in \verb|\boxed{}|.\\  
\\
Problem: \{\}
\end{tcolorbox}

%% file: prompt/prompt_hack.tex
\begin{tcolorbox}[colback=lightblue!50!white,colframe=lightblue,title=\textcolor{black}{The Prefixes Used in Prompt Hacking. Note that we list seven out of eleven languages.}, width=\textwidth, breakable]
\footnotesize
\label{prompt:prefix_used_prompt_hacking}
\begin{itemize}[left=0pt]
    % \item \textbf{English: }\texttt{<|Assistant|><think>\textbackslash n}By request, I will start thinking in English.
    % \begin{CJK}{UTF8}{min}
    % \item \textbf{Japanese: }\texttt{<|Assistant|><think>\textbackslash n}要求があれば、日本語で考え始めます。
    % \end{CJK}
    % \begin{CJK*}{UTF8}{gbsn} % This environment allows Chinese characters
    % \item \textbf{Chinese: }\texttt{<|Assistant|><think>\textbackslash n}应要求，我将开始用中文思考。
    % \end{CJK*}
    % \item \textbf{Spanish: }\texttt{<|Assistant|><think>\textbackslash n}A petición, empezaré a pensar en español.
    % \item \textbf{French: }\texttt{<|Assistant|><think>\textbackslash n}Sur demande, je commencerai à penser en français.
    % \item \textbf{German: }\texttt{<|Assistant|><think>\textbackslash n}Auf Anfrage werde ich anfangen, in Deutsch zu denken.
    % \item \textbf{Swahili: }\texttt{<|Assistant|><think>\textbackslash n}Kwa ombi, nitaanza kufikiria kwa Kiswahili.
    \item \textbf{English: }By request, I will start thinking in English.
    \begin{CJK}{UTF8}{min}
    \item \textbf{Japanese: }要求があれば、日本語で考え始めます。
    \end{CJK}
    \begin{CJK*}{UTF8}{gbsn} % This environment allows Chinese characters
    \item \textbf{Chinese: }应要求，我将开始用中文思考。
    \end{CJK*}
    \item \textbf{Spanish: }A petición, empezaré a pensar en español.
    \item \textbf{French: }Sur demande, je commencerai à penser en français.
    \item \textbf{German: }Auf Anfrage werde ich anfangen, in Deutsch zu denken.
    \item \textbf{Swahili: }Kwa ombi, nitaanza kufikiria kwa Kiswahili.
\end{itemize}
\end{tcolorbox}

%% file: prompt/template_for_r1_like_reasoning.tex
\begin{tcolorbox}[colback=lightblue!50!white,colframe=lightblue,title=\textcolor{black}{The Template for R1-like Reasoning}, width=\textwidth, breakable]
\footnotesize
\label{prompt:template_for_r1_like_reasoning}
You are a helpful AI Assistant that provides well-reasoned and detailed responses. You first think about the reasoning process as an internal monologue and then provide the user with the answer. The final answer must be put in \verb|\boxed{}|. Respond in the following format: \verb|<think>\n...\n</think>\n<answer>\n...\n</answer>|
\end{tcolorbox}